\documentclass[manuscript,screen]{acmart}

\AtBeginDocument{%
  \providecommand\BibTeX{{%
    \normalfont B\kern-0.5em{\scshape i\kern-0.25em b}\kern-0.8em\TeX}}}

\setcopyright{acmcopyright}
\copyrightyear{20xx}
\acmYear{20xx}
\acmDOI{10.xxxx/xxxxxxx.xxxxxxx}

\usepackage{algorithm}
\usepackage{algpseudocode} 
\usepackage{amsthm} 
\usepackage{array} 
\usepackage{enumitem}
\usepackage{graphicx}
\usepackage{longtable,ltcaption} 
\usepackage{multirow}
\usepackage{tabto} 
\usepackage{threeparttablex} 

\usepackage{fancyhdr,lipsum}
\fancypagestyle{firstpage}{
  \fancyhead[C]{\small To appear at the $6^{th}$ Workshop on Formal Methods for ML-Enabled Autonomous Systems (FoMLAS 2023), Paris, France.}
  \fancyfoot[C]{\thepage}
}
\begin{document}
\title{Scaling Model Checking for DNN Analysis via State-Space Reduction and Input Segmentation (Extended Version)}

\author{Mahum Naseer}
\orcid{0000-0003-3096-9536}
\affiliation{
    \institution{Institute of Computer Engineering, Vienna University of Technology}
    \city{Vienna}
    \postcode{1040}
    \country{Austria}}
\email{mahum.naseer@tuwien.ac.at}

\author{Osman Hasan}
\orcid{0000-0003-2562-2669}
\affiliation{
    \institution{School of Electrical Engineering \& Computer Science (SEECS), National University of Sciences \& Technology (NUST)}
    \city{Islamabad}
    \country{Pakistan}}
\email{osman.hasan@seecs.nust.edu.pk}

\author{Muhammad Shafique}
\orcid{0000-0002-2607-8135}
      
\affiliation{
    \institution{Division of Engineering, New York University Abu Dhabi (NYUAD)}
    \city{Abu Dhabi}
    \country{United Arab Emirates}}
\email{muhammad.shafique@nyu.edu}  

\begin{abstract}

Owing to their remarkable learning capabilities and performance in real--world applications, the use of machine learning systems based on Neural Networks (NNs) has been continuously increasing. 
However, various case studies and empirical findings in the literature suggest that slight variations to NN inputs can lead to erroneous and undesirable NN behavior. 
This has led to considerable interest in their formal analysis, aiming to provide guarantees regarding a given NN's behavior. 
Existing frameworks provide robustness and/or safety guarantees for the trained NNs, using satisfiability solving and linear programming. 
We proposed FANNet, the first model checking--based framework for analyzing a broader range of NN properties. 
However, the state--space explosion associated with model checking entails a scalability problem, making the FANNet applicable only to small NNs. 
This work develops state--space reduction and input segmentation approaches, to improve the scalability and timing efficiency of formal NN analysis. 
Compared to the state--of--the--art FANNet, this enables our new model checking--based framework to reduce the verification's timing overhead by a factor of up to $8000$, making the framework applicable to NNs even with approximately $80$ times more network parameters. 
This in turn allows the analysis of NN safety properties using the new framework, in addition to all the NN properties already included with FANNet. 
The framework is shown to be efficiently able to analyze properties of NNs trained on healthcare datasets as well as the well--acknowledged ACAS Xu NNs. 

\end{abstract} 



\maketitle
\thispagestyle{firstpage}

\section{Introduction}

The continuous improvement of Machine Learning (ML) systems, often wielding Neural Networks (NNs), has lead to an ever-growing popularity of these systems in real--world applications. Owning to their efficient learning and re-learning capabilities, high classification accuracy for testing datasets, and adaptability via transfer learning, they often find their way to numerous classification and decision--making tasks. These 
include face identification \cite{face-id}, speech recognition \cite{speech}, anomaly detection \cite{anomaly-detection}, and even safety critical applications like autonomous driving \cite{auto-drive} and healthcare \cite{healthcare,covid}.

However, as observed in numerous recent works, the NNs deployed in such systems are rarely \textit{resilient}, i.e., they are extremely susceptible to misclassify or arrive at an unsafe output decision in the presence of slight input modifications \cite{AdvAttack,deepxplore}. These modifications can be in the form of imperceptible noise \cite{trisec,fadec}, adversarial cues \cite{typographic-attack}, backdoors \cite{backdoor-2020}, or   simply affine and photometric transformations \cite{deepxplore}. Earlier attempts to ensure correct functioning of these NNs involved empirical approaches, for instance using gradient based methods \cite{AdvAttack,UniversalPerturb} to identify the adversarial noise patterns that would lead the NN to misclassify benign inputs. Although such attempts provide evidence to the lack of resilience of NNs, they are insufficient to provide any guarantees regarding NNs' resilience in the case when no adversarial noise is found. 

To deal with the aforementioned problem, there has been a great interest towards the rigorous evaluation of NNs, using formal verification, in recent years \cite{MIT,marabou,BaB2020}. This usually involves checking resilience properties, like robustness and safety, of the NNs using Satisfiability (SAT) checking or Linear Programming (LP). However, the exploration of formal approaches beyond SAT and LP, to analyze wider variety of NN's properties, remains largely neglected.


To the best of our knowledge, our prior work on Formal Analysis of Neural Networks (FANNet) \cite{naseer2020fannet} was the first attempt to analyze NN using model checking. The framework was applicable for the verification and analysis of multiple NN properties namely: robustness under constrained noise bounds, noise tolerance, training bias and input node sensitivity. \textit{However, the FANNet framework provided limited scalability for formal analysis owing to the large state--transition system it generated, even for relatively small NNs. Hence, the applicability of the framework was limited to small NNs only.}

This work introduces FANNet$+$ \footnote{\url{https://github.com/Mahum123/FANNetPlus}}, an enhanced model checking--based formal analysis framework for NNs that overcomes the limitations of FANNet, and provides significant improvement over FANNet in terms of scalability, timing--efficiency and the scope of NN properties analyzed by the framework. In particular, \textbf{the novel contributions of this work are as follows}:

\begin{enumerate}
    \item Providing novel state--space reduction techniques to reduce the size of the NN's state--transition system (Section \ref{sec:ssr}).
    \item Providing coarse--grain and random input segmentation approaches, to split the input domain into more manageable sub--domains, to aid model checking (Section \ref{sec:inp}).
    \item Leveraging the framework for the automated collection of a large database of counterexamples, which assist in an improved analysis of the sensitivity of input nodes and the detection of training bias (Section \ref{sec:comp_fm}).
    \item Comparing the timing overhead of simulation--based testing, FANNet \cite{naseer2020fannet} and the proposed framework. The proposed framework reduces the timing--cost of model checking by a factor of up to $8000$ (Section \ref{subsec:time}).
    \item Making use of the input sub--domains to verify safety properties of NNs, in addition to robustness under constrained noise, noise tolerance, training bias and input node sensitivity properties (Section \ref{subsec:safe}). 
    \item Deploying the above techniques to demonstrate the applicability of the new framework on NN case studies with up to $80$ times more parameters than the ones used in FANNet, thereby illustrating better scalability and applicability to more complex networks (Sections \ref{sec:nn} and \ref{sec:res}). 
\end{enumerate}

\paragraph{\textbf{Paper Organization}}
The rest of the paper is organized as follows. Section \ref{sec:rw} provides an overview of the formal analysis approaches available in the literature for NNs. Section \ref{sec:pre} defines the basic NN and model checking concepts and formalism relevant to this paper. Section \ref{sec:fannet} gives a summary of the framework FANNet. Section \ref{sec:pm} provides an overview for our proposed framework FANNet$+$ for the formal NN analysis, while elaborating on the proposed optimizations to improve the scalability, timing--efficiency and scope of addressed NN properties of the framework. Section \ref{sec:nn} highlights the NNs and datasets used to demonstrate the applicability of our framework for the various NN properties. Section \ref{sec:res} presents the results of analysis for the given NNs, also comparing timing--overhead of testing, FANNet and FANNet$+$. Finally, Section \ref{sec:conclusion} concludes the paper.

\section{Related Work} \label{sec:rw}

The earliest attempt \cite{2001verification} to analyze correct NN behavior involved mapping an NN to a look--up table. However, the approach lacked the sophistication to allow the analysis of intricate NN properties. Recent works instead focus mainly on the use SAT solving and LP, which not only allow a better formal representation of NNs but also the analysis of NN properties like robustness and safety.

\subsection{Satisfiability solving--based approaches}
The SAT--based NN verification works by first translating NN and its properties in Conjunctive Normal Form (CNF), and then using automated SAT solvers to search for the satisfying inputs that lead to erroneous NN behavior \cite{DPLL,CDCL}. However, the complexity of the problem limits its widespread use in real--world NN: the theory of transcendental functions, sometimes used as the activation functions for NNs, is known to be undecidable \cite{undecidable-theories}, and the verification of NN even with piecewise linear activations is an NP--complete problem \cite{reluplex}.

Earlier works \cite{pulina2012,isat3} focused on the safety properties of single hidden layer NNs with logistic function as the activation. More popular SAT-based approaches include the ones involving layer by layer NN analysis \cite{z3} and leveraging simplex algorithm to provide better splitting heuristics for piecewise linear activations during NN verification \cite{reluplex,marabou}. Some of the recent works \cite{vmware,BNNfortiss,OBDD} focus on the verification of Binary Neural Networks (BNNs), which are shown to be more power--efficient for the real--world applications \cite{bnn}. BNNs reduce the complexity of the verification task over conventional NNs by using binary valued parameters instead of the real numbered values.

In addition, the use of multiple abstraction techniques has also been proposed to reduce the complexity of SAT--based NN verification. These include input clustering \cite{deepsafe}, merging of the neurons in formal NN model \cite{BNNfortiss,neuron-merging}, and iterative approximation of a non-linear activation function to a piecewise linear activation function \cite{cetar}. However, it must be noted that approaches involving the approximation of non-linear function into piecewise linear ones \cite{cetar} may lead to \textit{incompleteness} in verification results, i.e., the results may contain false positives.

\subsection{Linear programming--based approaches}
In contrast, the LP--based NN verification approaches translate the NN to a set of linear constraints. The property to be verified is in turn expressed as an objective function. This formulates an optimization problem \cite{LPbook} aimed at identifying the satisfying assignments to the linear constraints that lead to the minima/maxima of the objective, and hence constitute an input leading to an erroneous behavior of the NN.

Due to the inherent support of LP for  the linear constraints, these approaches are often popular for the verification of NNs with piecewise linear (for instance, the ReLU) activations \cite{bastani2016,lomuscio2017,Neurify}. Branch and bound heuristics \cite{reluval,Neurify,BaB2018,BaB2020} are also often deployed for an efficient implementation of LP. In addition, some works \cite{FORTISS_lp,dutta2018,lomuscio2017,lomuscio2020-venus,MIT} also propose the use of Big-M technique, where an indicator variable is added to the linear constraints to distinguish the different linear regions of the piecewise linear activation function for NN verification.

The popular abstraction approaches to reduce the complexity of NN verification, pertinent to LP, include: overapproximation of the input domain \cite{NNV,Imagestar,matlab,matlab-incomp}, approximation of the ReLU activation function \cite{ai2,DeepZ,DeepPoly}, and merging of the neurons in NN layers \cite{munich}. In addition to these, experiment with the iterative deployment of both LP and SAT solvers over an overapproximated input domain was also proposed \cite{ehlers-planet}, to reduce the reachable output, and subsequently check network robustness and safety properties. Again, similar to the case for SAT--based approaches, such abstractions may render the verification \textit{incomplete}.

\section{Preliminaries} \label{sec:pre}

This section describes the formalism of NN architecture and properties relevant to this paper. The basics of model checking, sufficient to understanding this work, are also provided. Interested readers may refer to \cite{MCbook} for more inclusive details on model checking.

\subsection{Neural Network Architecture}
\begin{figure}[ht]
	\includegraphics[width=\linewidth]{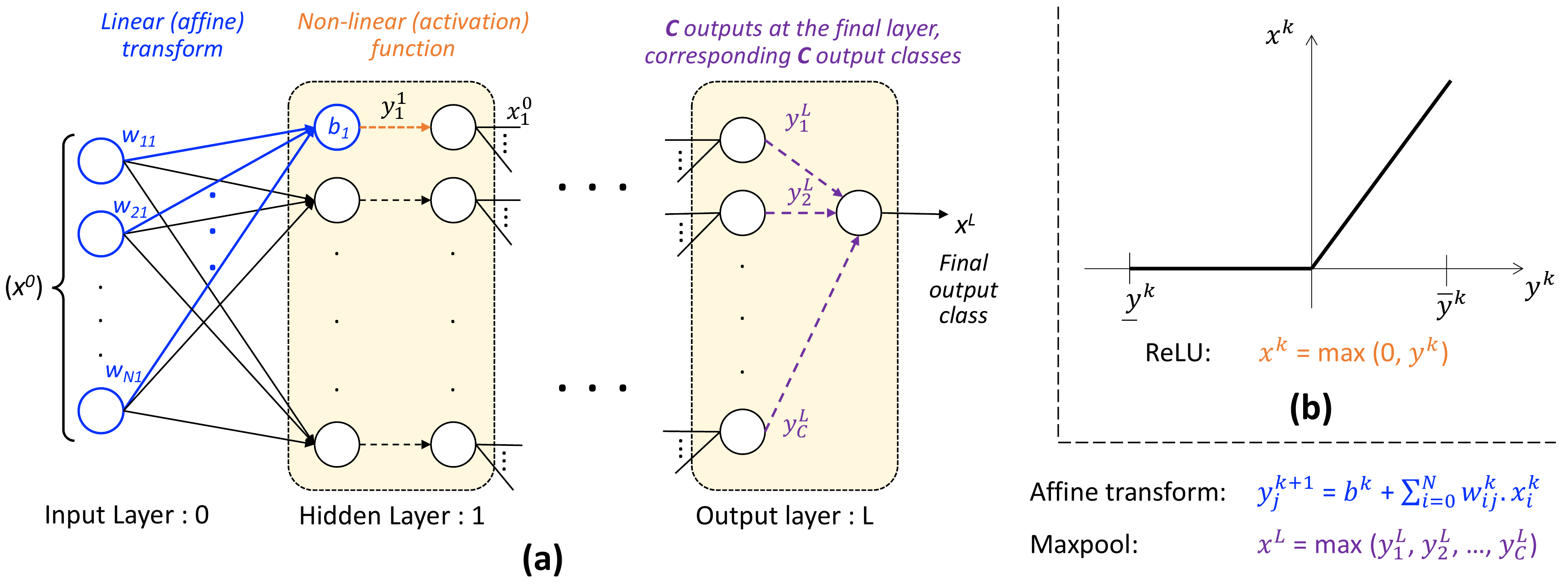}
	\caption{(a) A Feed--forward fully--connected neural network, with $L$ layers and $C$ output classes, and (b) the ReLU activation function.}
	\label{nn}
	\Description[]{Fully described in the text.}
\end{figure}

Essentially, a NN is an interconnection of neurons arranged in input, output and hidden layers \cite{ML-survey}. This work focuses on feed-forward fully-connected NNs, as shown in Fig. \ref{nn}(a). 

\begin{definition}[Feed--forward fully--connected neural network]
Given input domain $X^0$, a feed-forward network $F : X^0 \rightarrow X^L$ maps the input to the output domain $X^L$ such that the neurons in each layer $k$ depends only on the inputs from the preceding layer $k-1$. This results in a loop--free network that can be represented by $x^L = F~(x^0)~ = f^L~(~f^{L-1}~(...~f^1~(x^0)~...~))$, where $f^k$ encapsulates the linear and non--linear transformations for layer $k$, $x^0 \in X^0$ is an input from the input domain, and $x^L \in X^L$ is an output class from the $L$ classes in the output domain. The network is also fully--connected if each neuron in each network layer is connected to every neuron in the adjoining layer.
\end{definition}

Each layer of the network $F$ involves two transformations: a linear (affine) and a non--linear transformation. Given $N$ neurons/nodes in the layer $k$ of the network, the linear transformation can be given by:
\begin{equation}
    y^k_j = b_j^k + \sum_{i=0}^N w^k_{ij}x^{k-1}_i
\end{equation}
$w^k_{ij}$ represents the weight connecting neuron $i$ from layer $k-1$ to neuron $j$ in the layer $k$, while $b_j^k$ represents the bias parameter value corresponding neuron $j$ of the layer $k$.

The non--linear transformation maps output of the linear transformation via a non--linear activation function. For the purpose of this paper, all NN layers, except the output layer, use Rectified Linear Unit (ReLU) activation function. As shown in Fig. \ref{nn}(b), the ReLU is a piecewise linear function mapping negative inputs to zero, while using identity mapping the non--negative inputs:
\begin{equation}
    x^k = max(0,~y^k)
\end{equation}

The choice of NN output is often based on the output with the highest values. As such, maxpool function is used as an activation function for layer $L$ of the network:
\begin{equation}
    x^L = max(y^L_1,~y^L_2,~...,~y^L_C)
\end{equation}

\begin{figure}[ht]
	\includegraphics[width=\linewidth]{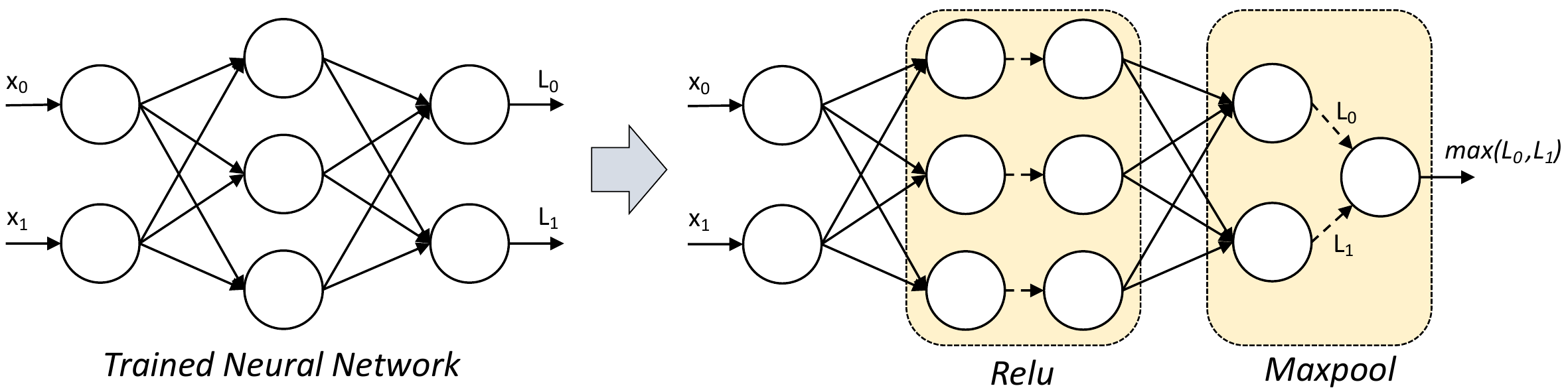}
	\caption{A single hidden layer fully--connected NN using ReLU and maxpool activation functions.}
	\label{run_eg}
	\Description[]{Running example for elaborating on the concepts described in preliminary section.}
\end{figure}

Consider the small feed--forward fully--connected NN shown in Fig. \ref{run_eg}. The input, hidden and output layers comprise of $2$, $3$ and $2$ neurons, respectively. The hidden layer uses ReLU activation function while the output layer makes use of maxpool function to provide the network's decision. The aforementioned network will be used as the running example in rest of the paper to visualize critical concepts. 

\subsection{Neural Network Properties}
From the literature, it is known that the output classification of the NN may be incorrect under certain input conditions. The following provides formalism to some essential NN properties to ensure correct NN behavior under varying input conditions (like the incidence of noise).

\begin{definition}[Robustness]
Given a network $F : X \rightarrow Y$, $F$ is said to be robust against the small noise $\Delta x$ if the application of the noise to an arbitrary input $x \in X$ does not change the output classification of input by the network, i.e., $\forall \eta \leq \Delta x:  F(x+\eta) = F(x)$.
\end{definition}
Hence, by definition, a robust NN does not misclassify inputs in the presence of pre--determined noise $\Delta x$.

\begin{definition}[Noise Tolerance]
Given a network $F  : X \rightarrow Y$, $F$ is said to be have a noise tolerance of $\Delta x_{max}$ if the application of any noise up to $\Delta x_{max}$ to an arbitrary input $x \in X$ does not change the output classification of input by the network, i.e., $\forall \eta \leq \Delta x_{max}:  F(x+\eta) = F(x)$.
\end{definition}

\noindent \textit{\textbf{Robustness versus Noise Tolerance --}} 
A NN with a noise tolerance of $\Delta x_{max}$ is robust against any noise $\Delta x$ such that $\Delta x \leq \Delta x_{max}$. Hence, noise tolerance is a stronger property compared to robustness. In other words, noise tolerance provides (an estimate of) the upper bound of the noise that the network $F$ can withstand, without showing any discrepancies in its normal behavior, i.e., without compromising the robustness of the network. 

\begin{definition}[Training/Robustness Bias]
Let $x_1 \in X$ be an arbitrary input from the output class $A \subset Y$, while $x_2 \in X$ be an arbitrary input from the output class $B \subset Y$. Given a network $F  : X \rightarrow Y$, $F$ is said to be biased towards the class $A$ if the addition of noise $\Delta x$ to $x_1$ does not cause any misclassification, but the addition of same noise to $x_2$ causes misclassification (i.e., $\exists \eta \leq \Delta x:  (F(x_1+\eta) = A) \land (F(x_2+\eta) \neq B)$), for a significantly large number of noise patterns ($\eta$).
\end{definition}

\noindent \textit{\textbf{Robustness versus Training Bias --}} 
In simple terms, training bias is the robustness of the individual output classes of the trained network, to the same magnitude of noise. Such biased behavior in NNs can often be a result of the use of imbalanced datasets \cite{toolboximbalance, imbalancedata2021}, i.e., datasets with significant portion of the input samples belonging to a particular output class (class $A$), during NN training. Hence, the resulting network is robust to bounded noise applied only to the inputs from class $A$.

\begin{definition}[Input Node Sensitivity]
Given a network $F  : X \rightarrow Y$, with the input domain comprising of $N$ nodes $X=[X_1,X_2,...,X_N]$. Let $\Delta x = [\Delta x_1,\Delta x_2,...,\Delta x_N]$ be the noise applied to an arbitrary input $x \in X$, where the noise applied to each node is within the same bounds, i.e., $\Delta x_1=\Delta x_2=...=\Delta x_N$. The input node $\alpha$ is said to be insensitive if the application of noise to the node $\forall \eta_\alpha \leq \Delta x_\alpha: x_\alpha+\eta_\alpha$ does not change the output classification of the input $x$. 
\end{definition}
It must be noted that the sensitivity of an input node is analyzed independent of the remaining input nodes. This means, if the application of noise $\Delta x_\alpha$ to the node $x_a$ does not change output classification of input $x$, the node is said to be insensitive regardless of what the noise (within the limits $\Delta x$) applied to the other input nodes may be.

\begin{definition}[Safety]
Given a network $F  : X \rightarrow Y$, let $[\underline{X}, \overline{X}] \subseteq X$ be the valid input domain, and $[\underline{Y}, \overline{Y}] \subseteq Y$ be the safe output domain. The network $F$ is said to be safe if all inputs within the valid domain maps to the safe output domain, i.e., $\forall x \in [\underline{X}, \overline{X}].~ F(x) = y ~s.t.~ y \in [\underline{Y}, \overline{Y}]$.
\end{definition}
The safety property can also be subsumed by the concept of reachability, which requires any undesired output to be unreachable by all valid inputs. It must also be noted that while considering safety property, the maxpool activation in the output layer of NN is omitted to allow the analysis of a continuous output domain $[\underline{Y}, \overline{Y}]$, as opposed to discrete output classes.


\subsection{Model Checking}
\begin{figure}[ht]
	\centering
	\includegraphics[width=\linewidth]{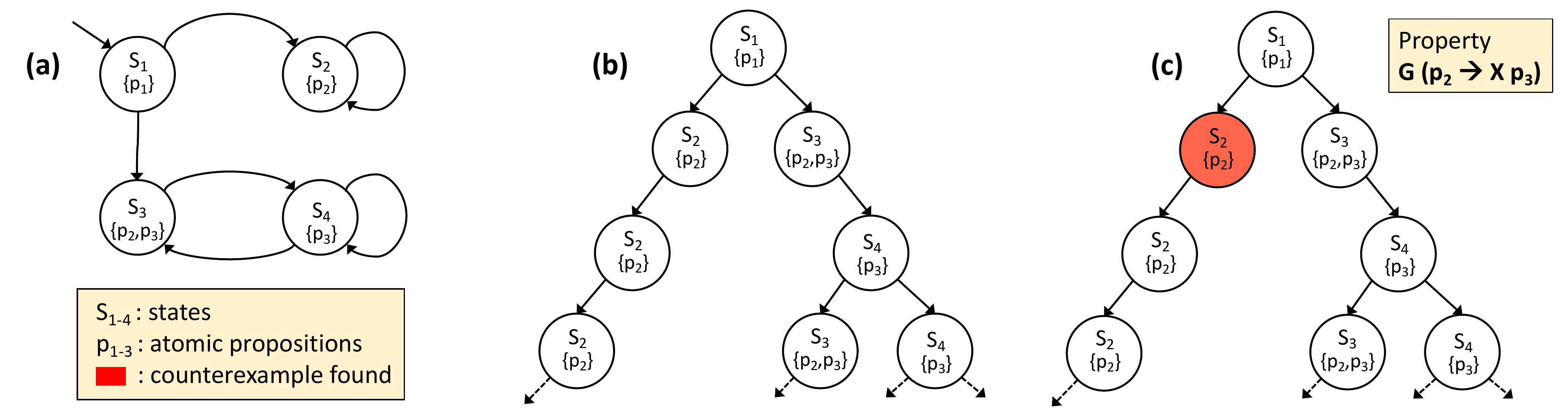}
	\caption{(a) A simple finite state transition system with four states and three atomic propositions of interest; (b) Unwinding the transition system to generate the computation tree for bounded model checking; (c) searching the computation tree for the temporal property \texttt{G} $(p_2 \implies$ \texttt{X} $p_3)$}  
	\label{MC}
	\Description[Example for BMC]{Given a state transition system with loop in state $S_2$, i.e., there is a transition from state $S_2$ to itself. Let $L(S_2)={p_2}$. Unwinding this system will generate a path ($S_2,S_2,...$) in the computation tree such that once the system enters the path, the path can not be left. Hence, the temporal property \texttt{G} $(p_2 \implies$ \texttt{X} $p_3)$ does not hold.}
\end{figure}

Model checking is an automated formal verification approach, whereby specifications/desired properties are rigorously checked for a formal model/implementation given as a state--transition system, like a Kripke structure, as shown in Fig. \ref{MC}(a). 

\begin{definition} [Kripke Structure]
Let $AP$ be the set of all possible atomic propositions for a given system. The Kripke structure $M$ for the system is then a tuple $M = (S,I,\delta,L)$ such that:
\begin{itemize}
    \item [--] $S$ is the set of all the possible states in the formal model $M$,
    \item [--] $I \subseteq S$ is the set of the possible initial states,
    \item [--] $\delta \subseteq S ~X~ S$ is the transition relation between the states, and
    \item [--] $L: S \rightarrow 2^{AP}$ is the labeling function that defines the $AP$ valid for each state in $M$.
\end{itemize}
\end{definition}

In case the specification does not hold for the formal model, the model checker generally provides the counterexample in the form of a path to property violation. 
The path $\rho$ in the model $M$ is the sequence of states $\rho = S_i, S_{i+1}, ...$ such that $\forall i. \delta(S_i,S_{i+1})$ holds. 
The paths may be \textit{finite}, i.e., comprising of a fixed number of states, or \textit{infinite}. 
The infinite paths result from self-loops or cycles in the model, and may contribute to the large number of states in the model, and hence, leads to the infamous \textit{state--space explosion} problem \cite{MCbook}. 
Additionally, the explicit declaration of state variables (as in explicit state model checking \cite{explicit}) may also lead to the generation of a large number states to encapsulate all possible behaviors in the formal model. 
This again leads to the state--space explosion.

\begin{figure}[ht]
	\centering
	\includegraphics[width=\linewidth]{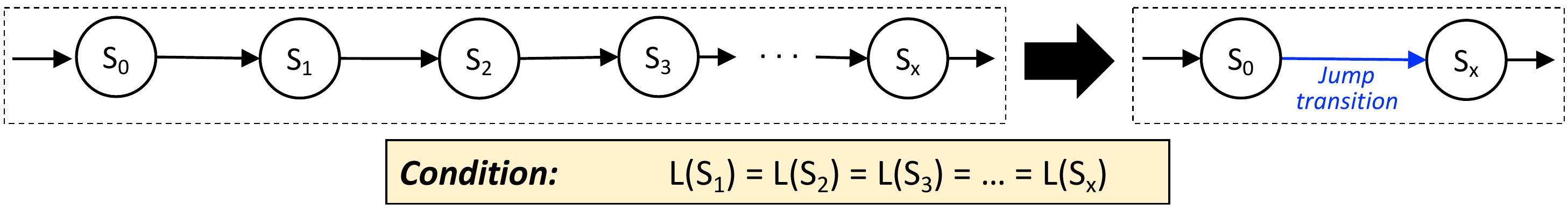}
	\caption{Compressing Kripke structure using jump transition.}  
	\label{trans}
	\Description[Jump transition reduces the number of states in Kripke Structure]{Given a path in the model through state $S_0,S_1,...S_x$ with all states having identical set of valid $AP$, i.e., $L(S_0)=L(S_1)=...L(S_x)$, then the path can be compressed to a shorter path with only states $S_0$ and $S_x$, and a single transition between these states.}
\end{figure}

Numerous abstraction approaches are available in the model checking literature to reduce the size of the formal model, hence avoiding state--space explosion and scaling the model checking for larger systems. Among the popular ones include: partial order reduction \cite{MCbook}, counterexample guided abstraction refinement (CEGAR) \cite{cegar}, and bisimulation \cite{bisimulation}. This paper leverages Bounded Model Checking (BMC) \cite{bmc}, Symbolic Model Checking (SMC) \cite{smc} and jump transitions \cite{jumpTransition} to optimize the formal models of NNs. BMC limits the length of paths by reducing number of reachable states, 
as explained later in the section. SMC provides a means to check system specifications without explicitly using a large number of states in the formal model by declaring the state variables symbolically, rather than explicitly. 
Jump transitions, on the other hand, compresses the states in the model for which the labeling function $L$ provides the same valid set of $AP$, as depicted in Fig. \ref{trans}. 


Temporal logic is often the preferred formalism for defining the system properties in model checking. It allows the notion of time to be expressed in the propositions, using temporal operators like:
\begin{itemize}
    \item [--] \textit{Next} (\texttt{X}): \texttt{X}$\phi$ holds true iff $\phi$ holds true in the next state.
    \item [--] \textit{Globally} (\texttt{G}): \texttt{G}$\phi$ holds true iff $\phi$ holds true in every state.
    \item [--] \textit{Eventually} (\texttt{F}): \texttt{F}$\phi$ holds true iff $\phi$ holds true in the current or (at least one of) the following states.
\end{itemize}

Given the formal model $M$ and a property defined in temporal logic, model checking involves unwinding the model (see Fig. \ref{MC}(b)) into a computation tree and transversing through the tree to search for a violation of the property (as shown in Fig. \ref{MC}(c)). Given loops in the model, the unwinding of the model can potentially lead to paths of infinite length, hence making model checking computationally infeasible. Hence, BMC is instead preferred, whereby the model is unwound a pre--defined number of times, so as to limit the maximum number of states on any given path in the tree by a bound $k$.

~\\
\noindent \textit{\textbf{Verification versus Falsification --}} Falsification is an empirical approach to find evidence to disprove system properties \cite{falsification}. Unlike verification, where the objective is to ensure an absence of erroneous behavior in the formal model, the objective of falsification is to merely find an evidence/counterexample presenting the erroneous behavior of the system's model.

In terms of NNs, the falsification has been widely studied under the literature pertaining to adversarial inputs \cite{cleverhans,carlini2017towards,khalid2021survey,aldahdooh2022survey}. NNs are already known to be vulnerable to noise \cite{AdvAttack}, hence providing misclassification under the incidence of even minute magnitudes of noise. Falsification utilizes various optimization techniques to search region around the seed inputs (to the NNs) to identify this kind of misclassifying noise, i.e., the adversarial inputs. However, the techniques often lack the formal reasoning to conclude an absence of adversarial inputs in case one is not found.
On the other hand, verification approaches rely on formal models and logical reasoning to ensure desired properties hold for the NNs. This makes them intrinsically different from the falsification techniques, albeit verification approaches like satisfiability solving and model checking also provide counterexample in case of property violation.
\section{Formal Analysis of Neural Networks (FANNet)} \label{sec:fannet}

This section provides an overview of our base framework FANNet \cite{naseer2020fannet} that employs a model checking--based approach for the formal analysis of neural networks. 
The framework uses model checking to obtain the noise tolerance of the trained NNs.
In addition, the counterexamples obtained from model checking are also leveraged to estimate sensitivity of the individual input nodes as well as detect any training bias in the trained NN.

\subsection{Generating Kripke Structure of the Neural Networks}
\begin{figure}[ht]
	\includegraphics[width=\linewidth]{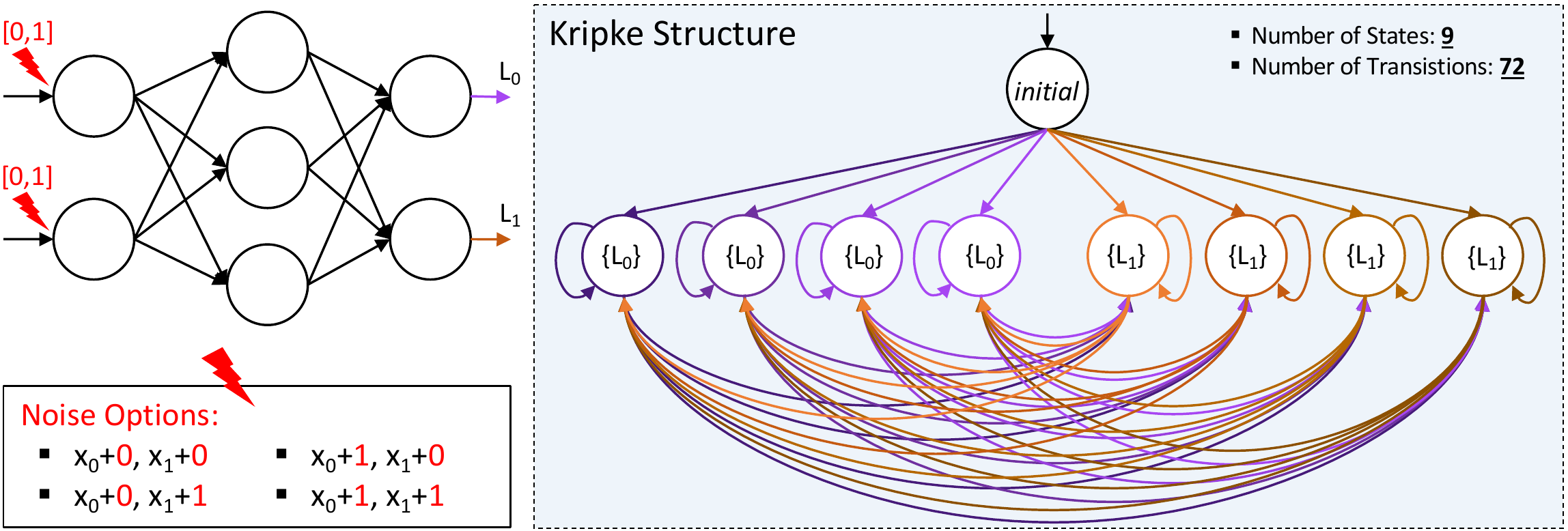}
	\caption{The formal model (Kripke structure) for a small fully--connected NN (shown on top left), as generated by FANNet. A binary noise is incident to the network's inputs.}
	\label{run_eg_fsm}
	\Description[]{Kripke structure for the trained NN.}
\end{figure}

Consider the NN from our running example (also shown in Fig. \ref{run_eg_fsm}). A binary noise, i.e., the noise with magnitudes either $0$ or $1$, is incident to the input nodes of the network. Recall that the model checking involves rigorous analysis of the state--transition system. So in case of explicit state model checking, there are four possible ways to apply noise to the input nodes. Hence, the resulting kripke structure contains four states leading from the initial state. 

Additionally, the input can lead to either of the possible outputs, which provide the atomic propositions ($AP$) for the formal model. Hence, a transition from initial state, corresponding to any of the possible noise options, can lead to any possible outputs. This leads to the resulting formal model (i.e., the Kripke structure) to have $1+nC$ states, where $n$ is the number of noise options and $C$ is the number of output classes in the network. Again, a rigorous analysis entails the possibility of network changing its state. Hence, the number of transitions making up the model adds to $nC(nC+1)$.
It must be noted that the hidden neurons are not presented in the formal model. This can be attributed to the use of jump transitions, which allows the compression of states having same set of $AP$. 

\begin{figure}[ht]
	\centering
	\includegraphics[width=\linewidth]{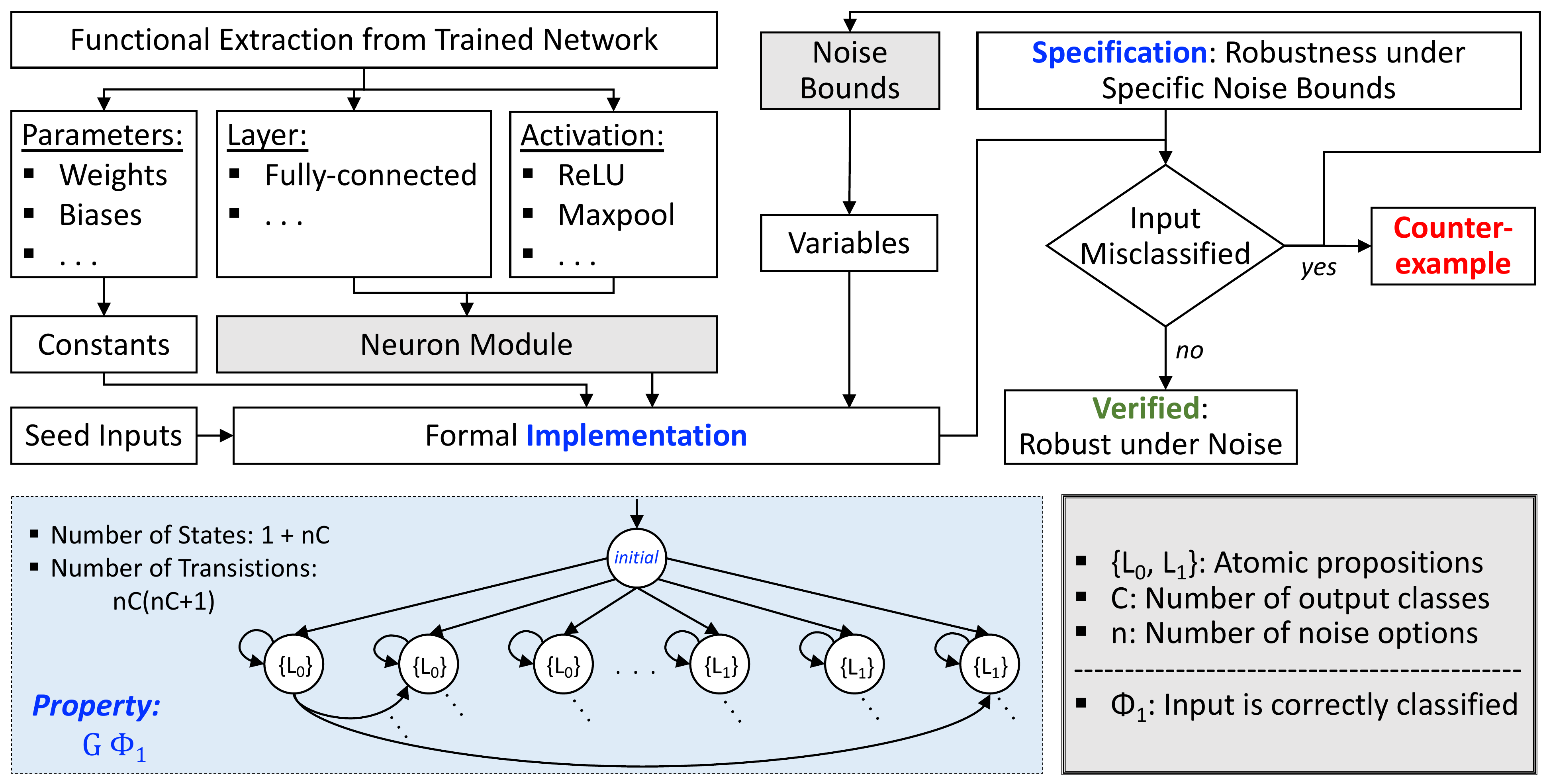}
	\caption{The flowchart presenting the use of model checking for checking NN noise tolerance, as proposed in FANNet. The blue box indicates the Kripke structure generated by the model checker for NN with two ($= C$) output classes/atomic propositions.}
	\label{fannet}
	\Description[Applying noise to network inputs results in large Kripke structure for the network]{Let $n$ be the number of noise options available at input and $C$ be the umber of output classes, the Kripke structure generated by FANNet comprises of ($1 + nC$) states and ($nC (1 + nC)$) transitions.}
\end{figure}

\subsection{Analyzing the Formal Network Model}
Initially, the parameters and architectural details of the trained NN, as well as the bounds of the applied noise to network inputs, are used to define the formal model of the network (as described earlier). 
Defining parts of the model in the modular form (for instance, using neuron module in Fig. \ref{fannet}) allows concise high-level description of the model to be written in the model checker.
The transitions between neurons (except the neurons in output layer) of the original NN are irrelevant to providing a change in output classification, i.e., for all paths leading up to the states representing neurons in output layers in the NN's model, the labeling provides same set of $AP$. 
Hence, these transitions can be compressed in the formal model using jump transitions, resulting in the model shown in Fig. \ref{fannet}. 

The model checker then checks the robustness of the NN under the application of noise within the defined noise bounds. 
The framework considers the model with a single seed input at a time. 
Therefore, once the output is classified, the class remains unchanged. 
Hence, the use of bounded model checking entails completeness while checking the robustness of the NN model. 
The violation of the desired network specification is accompanied by a counterexample, i.e., the specific noise vector that leads to misclassification of the seed input by the NN. 
Generally, large noise bounds are chosen initially. 
The model checking is then repeated while iteratively reducing the noise bounds, until network specifications hold valid. 
These noise bounds, $\Delta x_{max}$, represent the largest noise at which NN does not misclassify the input, and hence defines the noise tolerance of the network.

In addition to obtaining the noise tolerance, the framework can also be used to collect counterexamples, i.e., the noise vectors leading to misclassifications by the network. 
This is achieved by updating the formal specification with the negation of previously obtained counterexamples. 
Obtaining a significantly less number of misclassifying noise vectors that lead to a change in output class from $A$ to other classes, than from other classes to class $A$, indicates a bias of the NN towards class $A$.
A fine--grained analysis of the individual elements of the noise vectors, in turn, reflects on the predominant noise affecting/not affecting individual input nodes, hence unraveling the sensitivity of the nodes. 


\section{Proposed Optimizations for Formal NN Analysis} \label{sec:pm}


FANNet was an initial model checking--based framework, relying on explicit state model checking for the analysis of NN properties. 
Given the efficient performance of explicit state model checking for software verification problems in literature \cite{explicitvs1,explicitvs2,explicitvs3}, and the independence of our NN models on any particular hardware technology/components, the choice of the model checking approach used in FANNet was a logical first--step in the domain of model checking--based NN analysis. 

\begin{figure}[ht]
	\centering
	\includegraphics[width=\linewidth]{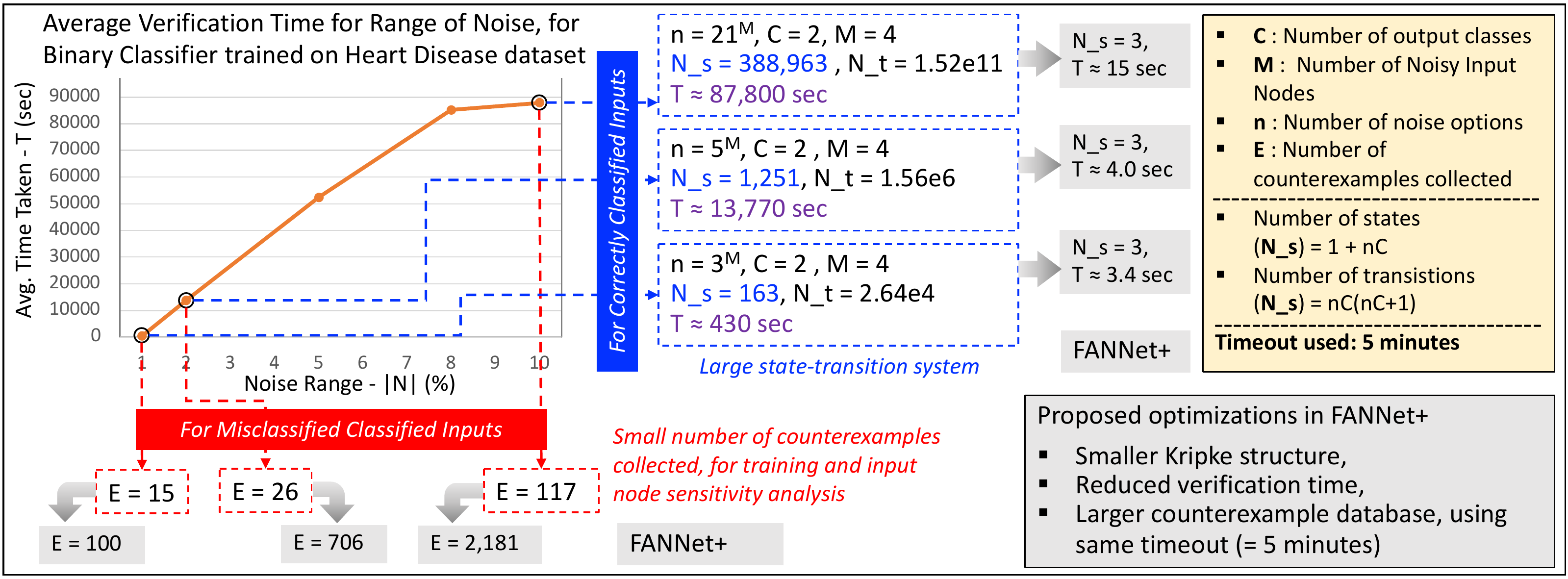}
	\caption{Formal analysis of a binary classifier trained on heart disease dataset using FANNet. The framework delineates limited scalability due to: (a) large size of the network's Kripke structure, (b) large verification time, and (c) small number of counterexamples collected within $5$ minutes timeout. The proposed optimizations for FANNet$+$ improve scalability of NN analysis by addressing all the aforementioned limitations.}
	\Description[Large difference in verification time, number of states in the network's Kripke structure, and size of counterexample database obtained using FANNet and FANNet$+$]{Unlike with FANNet$+$, where the number of states in Kripke structure stays constant, the the number of states using FANNet increases exponentially. Consequently, the verification time for FANNet is significantly larger as compared to FANNEt$+$. Given a constant timeout, this also leads to a larger number of counterexamples collected using our new framework.}
	\label{mot}
\end{figure}

However, as evident from the Kripke structure in Fig. \ref{fannet}, even a binary classifier will generate a formal model with $1+2n$ states, where $n$ depends on the size and precision of the noise bounds used. 
For instance, consider a binary classifier trained on heart disease dataset (details of the experimental setup and further results for the aforementioned classifier are provided in Sections \ref{sec:nn} and \ref{sec:res}). Fig. \ref{mot} shows the average verification time taken for varying input noise. As the noise bounds/range increase, the size of the resulting Kripke structure also increases. This accounts for the large timing overhead for the verification.

Moreover, for the precise analysis of NN's training bias and input node sensitivity, a large number of counterexamples is required. This means, the model checking needs to be repeated multiple times, while iteratively updating the specification of this large formal model. Again, it can be observed from the case study in Fig. \ref{mot} that running FANNet for a small time duration like $5$ minutes does not provide a large database of counterexamples for precise training bias and input node sensitivity analysis. These factors limit the scalability while increasing the timing overhead of FANNet.

This section elaborates on our proposed optimizations for reducing the size of the state--transition system (Kripke structure) leveraged by our enhanced framework, FANNet$+$. Moreover, two input splitting approaches are also proposed, which reduce the size of the input domain and hence improve the scalability and timing--efficiency of NN model. Hence, in addition to the multiple NN properties analyzed using FANNet, the proposed framework also allows the analysis of NN safety properties dealing with a large input domain, which was not viable earlier.

    
\subsection{State-Space Reduction} \label{sec:ssr}

As observed in the Kripke structure in Fig. \ref{fannet}, the state--space for the NN model contains multiple states leading to the same output class. This is because of the enumeration of the different noise combinations from the available noise bounds defining the formal model, leading to identical output states.

In contrast, FANNet$+$ proposes the use of SMC to reduce the identical output classes. The noise is added to the inputs symbolically, hence reducing the number of states in the model by a factor of approximately $n$. In terms of Kripke structure, the use of SMC to present noise symbolically is equivalent to merging of the states with the identical valid $AP$ in the NN model. 
In other words, if the transition relations $\delta(S_a,S_b)$ and $\delta(S_a,S_c)$ hold, and the labelling function defines identical $AP$ for the states $S_b$ and $S_c$, i.e., $L(S_b)=L(S_c)$, then the states $S_b$ and $S_c$ can be merged.

\begin{conjecture}
Given a model $M$ with $S=[S_a,S_b,S_c]$ and $\delta=[(S_a,S_b)$, $(S_b,S_b)$, $(S_a,S_c)$, $(S_c,S_c)]$ to be the set of all states and transition relations in the model, respectively, the states $S_b$ and $S_c$ can be merged iff $L(S_b)=L(S_c)$ holds.
\end{conjecture}

\begin{figure}[ht]
	\includegraphics[width=\linewidth]{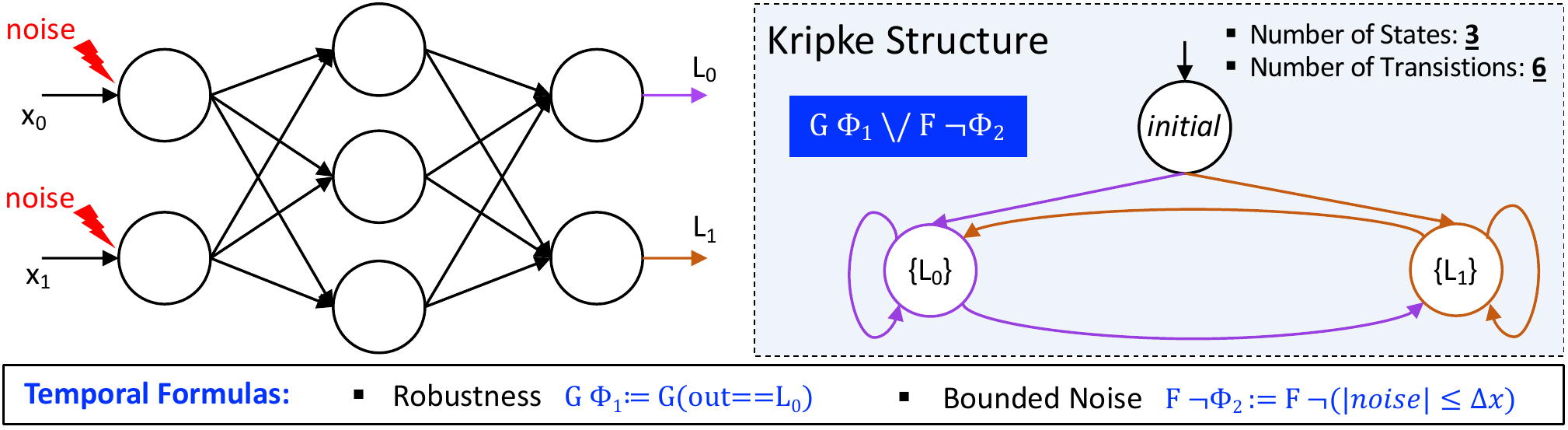}
	\caption{The formal model for a small fully--connected NN (shown on left), as generated by FANNet$+$, with noise added symbolically to the inputs.}
	\label{run_eg_redfsm}
	\Description[]{Kripke structure for the trained NN after state--space reduction.}
\end{figure}

To understand the above state--space reduction, consider again the network from our running example. As shown in Fig. \ref{run_eg_redfsm}, the noise is added to the inputs symbolically, i.e., $(x_0+noise)$ and $(x_1+noise)$. Hence, all the states with the same \textit{AP} (shown in Fig. \ref{run_eg_fsm}) can be merged. The resulting model is considerably smaller, containing only $1+C$ states and $C(C+1)$ transitions, where $C$ is the number of output classes.

It must be noted that by using SMC to reduce the state--space of the model, the noise bounds are instead added to the existing formal specifications of the NN model. This is evident in the running example, where the bounds to the noise are incorporated using \texttt{F}$\neg \Phi_2$. This is added to the existing property \texttt{G}$\Phi_1$. Together, the temporal property states that either the output is \textit{globally} correct or the noise \textit{eventually} exceeds the noise bounds. Naturally, this incurs additional complexity for checking the specification (i.e., the stated robustness property), but for a considerably less number of states in the model. Overall, this has a positive impact on both the scalability and the timing efficiency of the approach (as will be shown by our experimental results in Section: \ref{sec:res}).

Again from the Kripke structure in Fig. \ref{fannet}, it can be observed that the number of states with distinct valid $AP$ depends on the number of output classes of the NN. However, model checking provides binary answers while checking the specifications, i.e., either the specification holds (UNSAT) or it is violated (SAT). This allows the number of states to be reduced even further by considering the output of the NN to be either ``correctly classified" or ``misclassified" instead being one of the $C$--output classes of the network. The same reduction is also applicable while checking other NN properties, like safety.

\begin{figure}[ht]
	\centering
	\includegraphics[width=\linewidth]{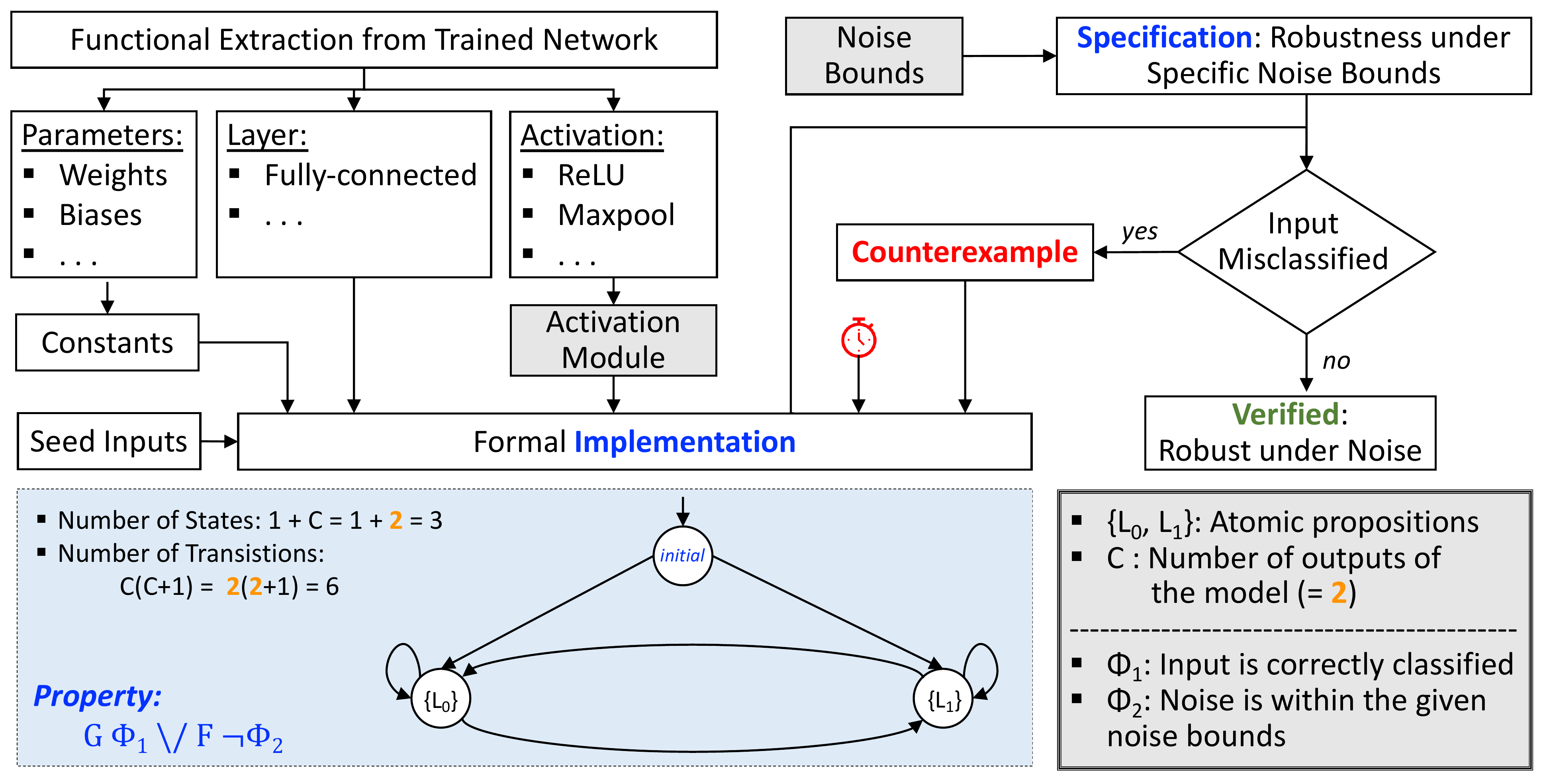}
	\caption{The flowchart presenting the use of symbolic model checking for checking NN noise tolerance. The blue box indicates the optimized Kripke structure generated by the model checker for the NN.}
	\Description[Providing noise as network specifications results in significantly smaller Kripke structure for the network]{Let $C$ be the umber of output classes, the Kripke structure generated by FANNet$+$ comprises of ($1 + C$) states and ($C (1 + C)$) transitions.}	
	\label{Novel}
\end{figure}

Overall, the model checking of the reduced state--space model is carried out as follows: NN parameters and architectural details are first used to define the NN formal model, as shown in the blue box in Fig. \ref{Novel}. A modular description of the model is again used for a concise high--level description of the model. The noise bounds and robustness property forms the specifications of the model. The model uses a fixed timeout to iteratively update the specification to collect counterexamples for the consequent analysis of NN properties, like training bias and input node sensitivity. The noise bounds are iteratively reduced until the noise tolerance of the trained network is obtained.  

\subsection{Input Domain Segmentation} \label{sec:inp}
Noise tolerance analysis, as described above, makes use of seed inputs. Hence, model checker only verifies specification for one element of the input domain at a time. However, for NN properties like safety, the verification often needs to be performed for the entire or subset of the input domain. This again requires a large Kripke structure, likely leading to state--space explosion. This paper proposes two approaches to resolve this problem: coarse--grain verification and random input segmentation.

\subsubsection{Coarse--grain Verification}
A rather straight--forward approach to verify a formal model with a large input domain is via sampling the input domain into discrete samples with regular intervals, i.e., with a constant sampling rate/step size. Depending on the size of the original NN, input domain and the available computational resources available to the model checker, the size of the input intervals can be fine--tuned. For NN specifications, for which the subset of input domain violating the specifications is large, the coarse--grain verification provides an efficient means to reduce the size of Kripke structure, while successfully finding any violations to the NN specifications. However, for input domains where property violation is a rare occurrence, the approach may overlook the property violations.

It must be noted that the use of SMC does not aid in the verification of systems with a large input domain, for highly non-linear systems like NNs. This is because, even though symbolic representation of input domain reduces the size of the generated Kripke structure, the addition of input bounds to the existing formal specification may make the specification too large to be verified by the model checker.

\subsubsection{Random Input Segmentation} \label{sec:ris}
\begin{figure}[ht]
	\centering
	\includegraphics[width=\linewidth]{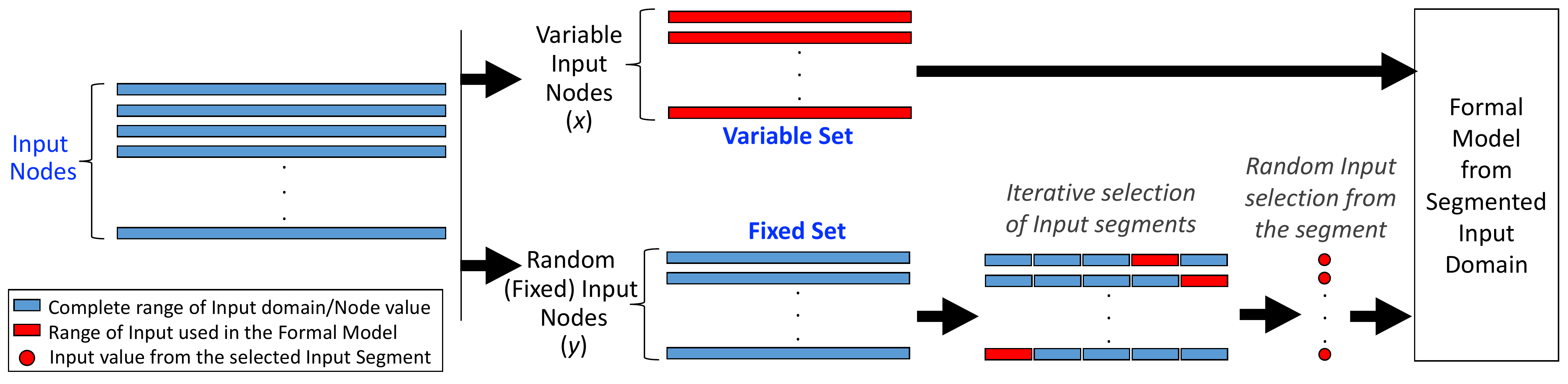}
	\caption{Overview of Random Input Segmentation: the input nodes are split into \textit{variable} and \textit{fixed} sets, which are then fed to the formal NN model.}
	\Description[]{Fully described in the text.}
	\label{seg}
\end{figure}
To address the challenge of dealing with a large input domain, this paper proposes the use of random input segmentation, as shown in Fig. \ref{seg}. The overall idea here is to divide the input nodes into two mutually exclusive sets: the \textit{variable} and the \textit{fixed} input node sets. The model checking is then carried out using the inputs from the \textit{variable} set represented symbolically while the discrete samples from the \textit{fix} set represented as constants in the model. The details for the approach, also highlighted in Algorithm \ref{algo1}, are as follows:
\begin{algorithm}[!t]{
			\footnotesize
			\caption{Random Input Segmentation}
			\label{algo1}
			\begin{algorithmic}[1]
				\Statex \textbf{Input:}\ \ \ \ \ Input Domain Bounds ($I$), Network Parameters ($w$, $b$, $L$, $N$), 
				\Statex \tab \ \ \ \ \ \ \ ~Normalization Parameters ($\mu$, $\varsigma$), Bins per Input Node ($X$), Specification ($\Phi$)
				\Statex \textbf{Output:}\ \ \ Counterexample ($CE$)
				\Statex \textbf{Initialize:} $CE \leftarrow [~]$
				\Statex
				
    \Statex //Creating bins to split ranges of each input node
    \For{$i=1$:Size($I,2$)} \Comment{For each input node}
    \For{$j=1$:$X(i)$} \Comment{For each input segment}
    \State $B[1][j][i] \leftarrow (\frac{I(2,i)-I(1,i)}{X(i)}\times(j-1)) + I(1,i)$
    \State $B[2][j][i] \leftarrow (\frac{I(2,i)-I(1,i)}{X(i)}\times j) + I(1,i)$
    \EndFor
    \EndFor
    
    \Statex //Segmentation
    \For{$i=1$:Size($I,2$)}
    \State $I' \leftarrow I$
    \State $Btemp \leftarrow B\setminus B\{:,:,i\}$
    \State $k \leftarrow $Size$(I,2) - 1$ \Comment{Total number of input nodes in \textit{fixed} set}

    \For{$j_1 = 1$:Size$(Btemp\{:,:,1\},2)$}
    \State ...
    \For{$j_k = 1$:Size$(Btemp\{:,:,k\},2)$} 
    
    \State $temp[1] \leftarrow rand(Btemp[1,j_1,1],Btemp[2,j_1,1])$
    \State ...
    \State $temp[k] \leftarrow rand(Btemp[1,j_k,k],Btemp[2,j_k,k])$ 
    
    \For{$m = 1:k$}  \Comment{For updating $I'$ with $temp$ for input nodes from \textit{fixed} set}
        \If {$i \leq m$}
				\State	$I'[m+1] \leftarrow temp[m]$
		\Else 
				\State	$I'[m] \leftarrow temp[m]$
		\EndIf
    \EndFor
    
    \State $temp \leftarrow [~]$
    \State $CE \leftarrow$ \textbf{FANN{\tiny ET}$+$}$(I',w,b,L,N,\mu,\varsigma,\Phi)$ \Comment{Model checking}
    
    \If{$CE \neq [~]$}
        \Return $1$ \Comment{Termination of code}
    \EndIf
    
    \EndFor
	\State ...
	\EndFor
	\EndFor
				
		\end{algorithmic}}
	\end{algorithm}
	

\begin{enumerate}
    \item Initially, for each input node, the upper and lower bounds of the (equally--spaced) input segments are calculated, as shown in Lines $3-4$ of Algorithm \ref{algo1}. The number of input segments for each node (i.e., the bins per input node ($X$)) is pre--defined.
    
    \item The input nodes for the \textit{variable} set are then picked, while the remaining nodes form the \textit{fixed} set. Line $9$ of Algorithm \ref{algo1} illustrates the selection of a single input node $i$ for the \textit{variable} set, while remaining nodes $Btemp$ form the \textit{fixed} set. However, the algorithm can be modified to work for any number of nodes being assigned to either of the sets. 
    
    \item Nested loops are used to pick the combination of segments for each input node in \textit{fixed} set. For each combination of segments, a random discrete input value is picked for each input node, as shown in Lines $11-16$ of Algorithm \ref{algo1}.
    
    \item The input domain is then updated with constant random values for input nodes in the \textit{fixed} set, as shown in Lines $17-23$ of Algorithm \ref{algo1}.
    
    \item The model checking is then performed with this updated input domain, shown by Line $25$ of Algorithm \ref{algo1}.
    
    \item If a counterexample corresponding to a property violation is obtained (as shown in Line $26$ of Algorithm \ref{algo1}), i.e., the property is found to be SAT, model checking can be terminated. However, if no counterexample is found, the algorithm proceeds with the next discrete sample from the combination of segments of input nodes from the \textit{fixed} set.
    
    \item In turn, the process is repeated with a new splitting of input nodes into \textit{variable} and \textit{fixed} sets, as depicted in Lines $7-9$ of Algorithm \ref{algo1}.
\end{enumerate}

As the number of input nodes fed to the NN increase, the computation requirements (and subsequently the timing overhead) also increase. 
This ``curse of dimensionality" is a known challenge with NN analysis in the literature \cite{survey2020}. 
However, by using random input segmentation, each splitting of the input nodes into the two sets and the following model checking are completely independent. 
This provides an opportunity for high degree of parallelism to the approach by dealing with different combination of nodes from \textit{fixed} and \textit{variable} sets using a different core. 
This, in turn, reduces the timing--overhead of the analysis.

It must be noted that RIS is optimal for DNN verification (using FANNet$+$) iff $I^n > M^n.\frac{I!}{M!M'!}$ hold, where $n$ is the number of noise options, $I$ is the total number of input nodes, $M$ is the number of nodes in the variable set and $M'$ the number of nodes in the fixed set ($I=M+M'$).

~\\
\noindent \textit{\textbf{Soundness -- }} As shown in Fig. \ref{seg} and Lines $17-23$ of the Algorithm \ref{algo1}, the input domain $I'$ contributing to the formal model analyzed by the model checker is the combination of the entire bounds of the input nodes from the \textit{variable} set and random inputs from the selected segments of input nodes from the \textit{fixed} set. Hence, this updated input domain is the proper subset of the original valid input domain $I$ of the NN, i.e., $I'\subset I$. This preserves soundness of the framework.

~\\
\noindent \textit{\textbf{Completeness -- }} Since the updated input domain $I'$ is the proper subset of domain $I$, the model checking using the domain entails incompleteness. This is the direct result of the Lines $14-16$ of Algorithm \ref{algo1}, which bypass the exhaustive coverage of the segments of the input nodes from the \textit{fixed} set. However, it must be noted that such incompleteness is essentially different from the incompleteness often observed in NN analysis literature \cite{DeepZ,matlab-incomp,DeepPoly,matlab} arising from overapproximation of the input domain and/or activation functions, and hence leading to false positives in the analysis. The analysis based on our algorithm, in contrast, does not lead to any false positives.

\subsection{FANNet\texorpdfstring{$+$} :: An Integrated and Optimized Framework for Formal NN Analysis} \label{sec:comp_fm}
\begin{figure}[ht]
	\centering
	\includegraphics[width=\linewidth]{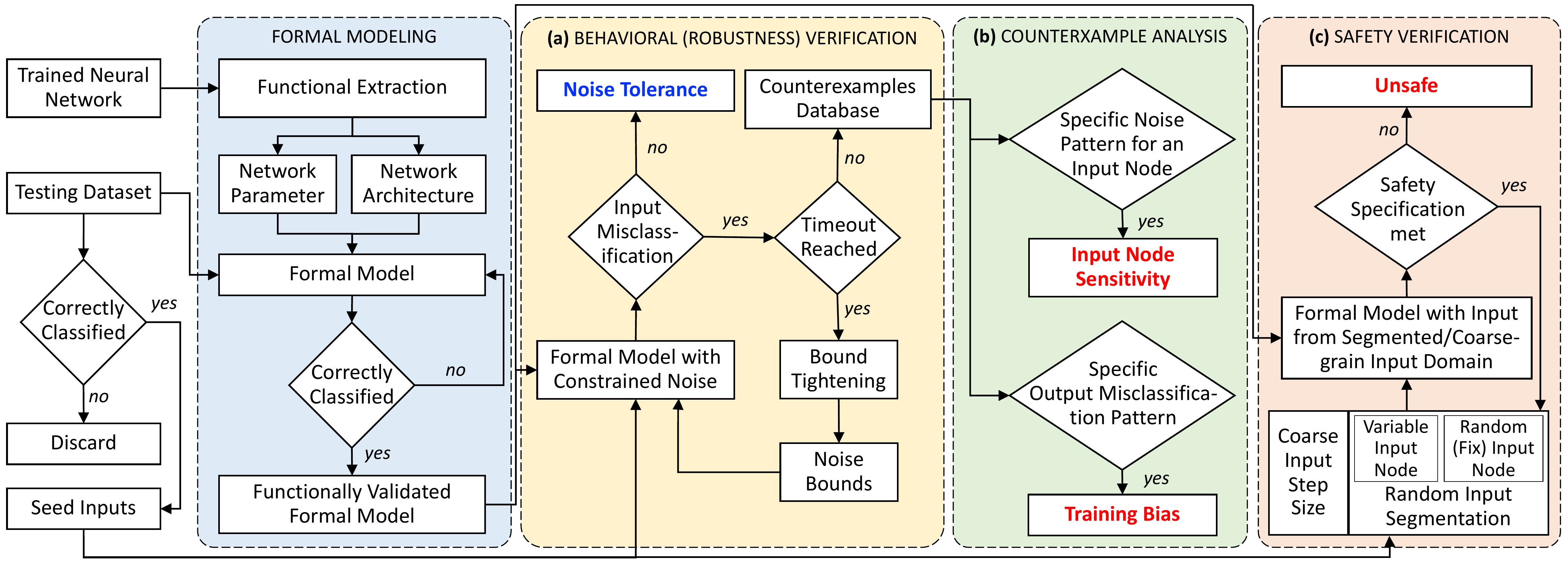}
	\caption{FANNet$+$: the proposed (optimized) framework for the formal analysis of trained fully-connected neural networks. The framework is able to: (a) provide robustness verification and ultimately obtain noise tolerance of the network, (b) use the obtained counterexamples to analyze sensitivity of individual input nodes and check for the presence of training bias, and (c) provide safety verification using coarse--grain and/or randomly segmented input domain.}
	\label{Meth}
	\Description[]{Fully described in the text.}	
\end{figure}

The proposed optimizations reduce the size of NN's Kripke structure as well as split the input domain into more manageable sub--domains. This improves the scalability and timing--efficiency of the framework, also allowing the analysis of NN properties beyond those addressed in FANNet. Fig. \ref{Meth} summarizes our overall proposed framework FANNet$+$ for formal NN analysis.

Initially, the formal model of the NN is defined in the appropriate syntax of the model checker, as indicated by the blue box in Fig. \ref{Meth} and Algorithm \ref{algo2}. This requires the use of trained NN parameters and architectural details of the network. Input is often normalized prior to being sent to the NN (as shown in Line $2$ of Algorithm \ref{algo2}). Some NNs may in turn also use inverse--normalization for the NN output (as shown in Line $14$ of Algorithm \ref{algo2}). To validate the functional correctness of the model, the output of the model is checked for the known (testing) inputs.

\begin{algorithm}[ht]{
			\footnotesize
			\caption{Network Formalization and Property Checking}
			\label{algo2}
			\begin{algorithmic}[1]
				\Statex \textbf{Input:}\ \ \ \ \ Input ($I$), Network Parameters ($w$, $b$, $L$, $N$), \Statex \tab \ \ \ \ \ \ \ Normalization Parameters ($\mu$, $\varsigma$), Specification ($\Phi$)
				\Statex \textbf{Output:}\ \ \ Counterexample ($CE$)
				\Statex \textbf{Initialize:} $CE \leftarrow [~]$
				\Statex
	
	\Function{FANNet$+$}{$I,w,b,L,N,\mu,\varsigma,\Phi$}
    
    \State $I_{norm} \leftarrow \frac{I-\mu}{\varsigma}$ \Comment{Input normalization}
    \State $temp\_I \leftarrow I_{norm}$
    
    \For{$i=1:L$} 
    \For{$j=1:N(i)$} 
        \State $temp\_O[j] \leftarrow \sum (w[j][:][i]) \times temp\_I[j]) + b[j][i])$ \Comment{Fully--connected layer}
    
        \If{$temp\_O[j] < 0$}
        \State $temp\_O[j] \leftarrow 0$ \Comment{ReLU activation}
        \EndIf
    \EndFor
    \State $temp\_I \leftarrow temp\_O$ \Comment{Input propagation to next NN layer}
    \EndFor

    \State $Out_{norm} \leftarrow temp\_O$
    \State $Out \leftarrow (Out_{norm} \times \varsigma) + \mu$ \Comment{Output inverse--normalization}
    
    \State $CE \leftarrow$ \texttt{Property}$(\Phi,I,Out)$ \Comment{Checking NN specification}
    \State \Return $CE$
	\EndFunction
				
		\end{algorithmic}}
	\end{algorithm}
	

The robustness verification of the validated formal model is carried out using seed inputs from the testing dataset and noise bounds, as explained in Section \ref{sec:ssr}. This is expressed in the yellow box in Fig. \ref{Meth}. Given the noise bounds $N\%$, the noise applied to seed inputs $\Delta x$ is taken to be a percentage of original input node values, i.e., $\Delta x = x * (N/100)$. Hence, the robustness of the model is checked within L$^\infty$--ball (i.e., infinity norm) of radius $\Delta x$ around the seed input. The NN specification is updated iteratively with the obtained counterexamples until a pre--defined timeout is reached. This allows the generation of a large counterexample database. In case the property holds before reaching the timeout, the model checking is immediately terminated. The noise bounds are iteratively reduced until the noise tolerance of the given network is obtained.

The counterexample database is then used in an empirical analysis to check the sensitivity of individual input nodes and detect any underlying training bias. The is shown by the green box in Fig. \ref{Meth}. The process is similar to the one used in FANNet. However, the improved timing--efficiency of our current framework allows model checking a large number of times within the pre--defined timeout. This provides a much larger counterexample database than was possible with FANNet, which in turn allows better analysis of the NN properties in question.

NN safety properties involve checking the NN model with a large input domain. The optimizations proposed in Section \ref{sec:ris} allow safety verification of the validated formal model, as shown in the orange box in Fig. \ref{Meth}. Here, either coarse--grain verification or verification with random input segmentation can be opted. As mentioned earlier, the sampling rate (in coarse--grain verification) and the size of input segments (in random input segmentation) is chosen on the basis of the size of original NN, input domain and the available computational resources available to the model checker. The larger the network, input domain or computation resources available, the higher the sampling rate (or the smaller the input segment) needs to be.

\section{Experiments} \label{sec:nn}

This section describes the datasets and the NNs trained on them, which are used to demonstrate the application of FANNet$+$ for the analysis of NN properties. 

\subsection{Leukemia Type Identification} 
The leukemia dataset \cite{leukemia-dataset} is a collection of $72$ samples of $7129$ genetic attributes of leukemia patients. The dataset is split into $38$ and $34$ training and testing samples, respectively. Based on the genetic attributes, the dataset categorizes two types of leukemia: Acute
Lymphoblast Leukemia (ALL), represented by approximately $70\%$ of the training samples, and Acute Myeloid Leukemia
(AML), represented by the remaining $30\%$ of the training samples. 

Minimum Redundancy and Maximum Relevance (mRMR) feature selection \cite{matlab-leukemia} was used to extract the $5$ most important genetic attributes representing leukemia, which were then used for the training. A feed--forward fully--connected NN with single hidden layer and a total of $141$ network parameters was trained for the dataset. The training and testing accuracies of the NN were $100\%$ and $94\%$, respectively.

The following properties are analyzed for this NN: robustness (under constrained noise), noise tolerance, input node sensitivity, and training bias. The results of the analysis are presented in Section \ref{sec:res}.

\subsection{Heart Disease Prognosis} 
The heart disease dataset \cite{heart-dataset} provides the records for $13$ attributes of the patients: among these, $4$ attributes are represented by continuous values while remaining represent discrete attributes. 
The outputs indicate if the patients' blood vessels are narrowing, hinting to a heart disease. The dataset is split into training and testing datasets with $261$ and $42$ inputs, respectively. Approximately $55\%$ of the samples from training dataset correspond to the case with narrowing blood vessels, while the remaining represent the samples with no significant narrowing of the blood vessels.

A feed--forward fully--connected NN with $3$ hidden layers, and a total of $622$ network parameters was trained for this dataset. The training and testing accuracies of the NNs were $90\%$ and $86\%$, respectively.

To imitate real--world case scenarios, where the noise is much more likely to affect continuous variables as compared to discrete ones, noise was applied to the input nodes with continuous variables. This was in turn used for the analysis of robustness (under constrained noise), noise tolerance, input node sensitivity, and training bias. The results of the analysis are presented in Section \ref{sec:res}.

\subsection{Airborne Collision Avoidance System (ACAS Xu)} 
Airborne Collision Avoidance System (ACAS Xu) NNs belongs to the family of ACAS X systems that make use of the trajectories of ownship and intruder to ensure safety while maneuvering the ownship. The system consists of $45$ NNs, which correspond to approximately different parts of the input domain. Each NN is feed--forward and fully--connected, with ReLU activation function, $6$ hidden layers and $13,305$ parameters.

Let $i_1$--$i_5$ be the NN inputs corresponding distance between ownship and intruder, and their heading angles and speeds, while $o_1$--$o_5$ be the ownship's maneuvering decisions namely clear--of--conflict, weak left/right and strong left/right. The NN's output decision corresponds to the output class with minimal value. Then the safety properties relevant to the NNs, which are also well--studied in the literature, include:

\begin{enumerate}
    \item Given a large distance between ownship and intruder, with the intruder travelling at much slower speed than ownship, the clear--of--conflict remains below a certain threshold.
    \begin{equation*} \label{p1} \tag{P1}
        (i_1 \geq 55947.691)~ \land (i_4 \geq 1145) \land (i_5 \leq 60) \implies (o_1 \leq 1500)
    \end{equation*}

    \item Given a large distance between ownship and intruder, with the intruder travelling at much slower speed than ownship, the clear--of--conflict advisory does not have the highest value.
    \begin{equation*} \label{p2} \tag{P2}
        (i_1 \geq 55947.691) \land (i_4 \geq 1145) \land (i_5   \leq 60) \implies (o_2 > o_1) \lor ~(o_3 > o_1) \lor (o_4 > o_1) \lor ~(o_5 > o_1)
    \end{equation*}

    \item Given a constrained distance between ownship and intruder, with the intruder in line of ownship's translation and moving towards it, the NN's clear--of--conflict advisory is not minimal.
    \begin{equation*} \label{p3} \tag{P3}
        \begin{split}
            (1500 \leq i_1 \leq 1800) &\land (-0.06 \leq i_2 \leq 0.06) \land (i_3 \geq 3.10) \land i_4 \geq 980) \land (i_5 \geq 960) \\
           & \implies (o_2 < o_1) \lor (o_3 < o_1) \lor (o_4 < o_1) \lor (o_5 < o_1)
        \end{split}
    \end{equation*}
    
    \item  Given a constrained distance between ownship and intruder, with the intruder in line of ownship's translation and moving away from it, but with a speed slower than that of the ownship, the NN's clear--of--conflict advisory is not minimal.
    \begin{equation*} \label{p4} \tag{P4}
        \begin{split}
            (1500 \leq i_1 \leq 1800) &\land (-0.06 \leq i_2 \leq 0.06) \land (i_3 = 0) \land ~(i_4 \geq 1000)~ \land (700 \leq i_5 \leq 800) \\
            &  \implies (o_2 < o_1) \lor (o_3 < o_1) \lor (o_4 < o_1) \lor (o_5 < o_1)
        \end{split}    
    \end{equation*} 
\end{enumerate}

As indicated in Section \ref{sec:pm}, the analysis of safety properties involves a formal model with large input domains. Hence, coarse--grain verification and random input segmentation are used for the analysis of ACAS Xu NNs. 
\section{Results and Analysis} \label{sec:res}
We use NNs highlighted in the previous section to perform formal NN analysis on CentOS-$7$ systems running on Intel Core i$9-9900$X processors at $3.50$GHz. The proposed framework FANNet$+$ is implemented in Python, C++ and MATLAB, and uses NuXmv model checker \cite{nuxmv} back end. 
The tools Reluplex \cite{reluplex} and Marabou \cite{marabou} are implemented on virtualbox running Ubuntu $18.04$, for comparison. The timeout used for NNs trained on leukemia and heart disease datasets is $5$ minutes, while a timeout of $2$ hours is used for ACAS Xu NNs. 

This section first compares the timing and memory overhead of simulation--based testing (using MATLAB) with the FANNet. The timing performance of FANNet is then compared to FANNet$+$, for robustness, noise tolerance, training bias and input node sensitivity analysis. Finally, the safety verification results for ACAS Xu NNs are presented, while comparing these results to those obtained from popular NN verification tools Reluplex and Marabou. 
It must be noted that the objective of comparison between FANNet$+$ and SMT-based tools is to establish the consistency of results, not to compare timing-overhead since model checking is still fairly a new direction for DNN analysis.

\subsection{Comparative Analysis of the Computational Overhead for Testing and FANNet} \label{subsec:time}
Testing is generally considered more user friendly, as compared to model checking. However, the results for model checking are more rigorous, and hence provide more reliable behavioral guarantees than testing. We compared the performance of FANNet with MATLAB--based testing. For testing, we define a matrix for all possible noise combinations, for a predefined noise bounds, before initializing the test. On the other hand, the model checker searches for noise combinations, non-deterministically, at run--time. Both experiments are based on the small NN trained on the Leukemia dataset, as described in the previous section, since the timing and memory overhead of testing increases rapidly for large noise bounds for larger NNs. Both experiments run on the same seed inputs. 

\begin{figure}[ht]
	\centering
	\includegraphics[width=\linewidth]{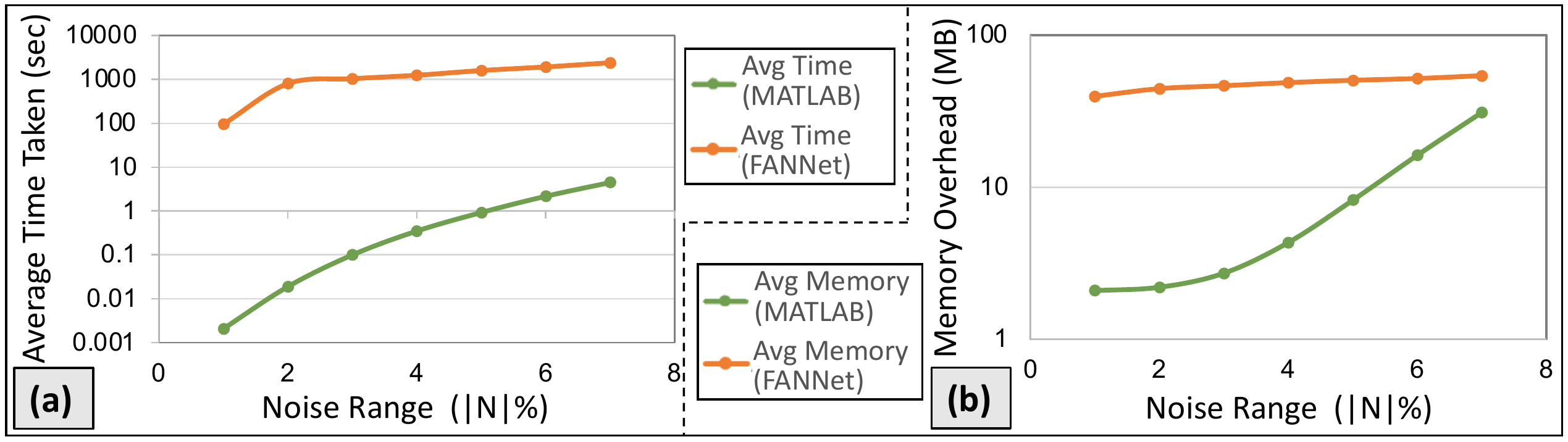}
	\caption{(a) Timing, and (b) Memory overhead comparison, of analyzing the neural network trained on Leukemia dataset, between exhaustive testing (using MATLAB) and the model checking--based framework FANNet \cite{naseer2020fannet}.}
	\label{FANNetvsMATLAB}
	\Description[]{Fully described in the text.}
\end{figure}

Considering the time taken until the termination of both experiments, the average timing requirement of the FANNet, although significantly higher than testing's for the given experiment, increases at a slower rate than that for testing. This trend is illustrated in Fig.  \ref{FANNetvsMATLAB}(a). On the other hand, the increase in average memory requirements of FANNet also increases at a significantly slower rate than testing, as shown in Fig.  \ref{FANNetvsMATLAB}(b). 

The trends for the average time and memory requirements of both experiments indicate that the strength of model checking is more prominent for larger NNs. As the size of the network grows, the Kripke structure system for NN grows, resulting in the increased latency and utilized memory resources. On the other hand, the computational requirements for (exhaustive) testing also increase with the increase in input noise range, often at a much higher rate than that for FANNet. This establishes the \textit{advantage of formal NN analysis, even when the framework is not optimized}.

\subsection{Behavioral (Robustness) Verification and Noise Tolerance Determination}
\begin{table}[]
\caption{Robustness of seed inputs from leukemia dataset under varying noise bounds}
\label{tab:leu}
\begin{tabular}{lrrrrl|lrrrr}
\toprule
\multicolumn{1}{c}{\multirow{2}{*}{\textbf{Input}}} & \multicolumn{4}{c}{\textbf{Result with   Noise:}}         &  & \multicolumn{1}{c}{\multirow{2}{*}{\textbf{Input}}} & \multicolumn{4}{c}{\textbf{Result with   Noise:}}  \\
\multicolumn{1}{c}{} & \multicolumn{1}{l}{$11\%$} & \multicolumn{1}{l}{$20\%$} & \multicolumn{1}{l}{$30\%$} & \multicolumn{1}{l}{$40\%$} &  & \multicolumn{1}{c}{} & \multicolumn{1}{l}{$11\%$} & \multicolumn{1}{l}{$20\%$} & \multicolumn{1}{l}{$30\%$} & \multicolumn{1}{l}{$40\%$} \\
\midrule
1   & UNSAT     & UNSAT     & UNSAT  & UNSAT  &  & 17       & UNSAT    & UNSAT   & UNSAT  & UNSAT   \\
2     & UNSAT   & UNSAT  & UNSAT  & UNSAT &  & 18          & UNSAT   & UNSAT   & UNSAT & UNSAT   \\
3    & UNSAT   & UNSAT  & UNSAT   & UNSAT &  & 19          & UNSAT   & UNSAT   & UNSAT   & UNSAT   \\
4   & UNSAT    & UNSAT   & UNSAT  & UNSAT &  & 20          & UNSAT     & SAT    & SAT  & SAT    \\
5    & UNSAT  & UNSAT  & UNSAT  & UNSAT  &  & 21            & UNSAT   & UNSAT   & UNSAT    & UNSAT  \\
6  & UNSAT  & UNSAT  & UNSAT  & UNSAT  &  & 22             & UNSAT    & UNSAT   & UNSAT & UNSAT   \\
7  & UNSAT & UNSAT  & UNSAT   & UNSAT   &  & 23             & UNSAT    & UNSAT  & UNSAT  & UNSAT   \\
8  & UNSAT  & UNSAT  & UNSAT  & UNSAT  &  & 24              & UNSAT   & UNSAT  & UNSAT  & SAT   \\
9   & UNSAT  & UNSAT   & UNSAT  & UNSAT  &  & 25            & UNSAT   & UNSAT   & SAT   & SAT   \\
10  & UNSAT  & UNSAT  & UNSAT    & UNSAT      &  & 26        & UNSAT   & UNSAT   & UNSAT   & UNSAT  \\
11 & UNSAT  & UNSAT   & UNSAT & UNSAT &  & 27                & UNSAT    & SAT  & SAT  & SAT     \\
12   & UNSAT  & UNSAT   & UNSAT  & UNSAT  &  & 28           & UNSAT  & UNSAT   & UNSAT   & UNSAT   \\
13  & UNSAT  & UNSAT  & UNSAT   & UNSAT  &  & 29           & UNSAT  & UNSAT   & SAT  & SAT    \\
14  & UNSAT  & UNSAT  & UNSAT  & UNSAT  &  & 30             & UNSAT   & SAT  & SAT & SAT     \\
15  & UNSAT  & UNSAT  & UNSAT  & SAT &  & 31                 & UNSAT  & UNSAT  & UNSAT  & UNSAT  \\
16 & UNSAT  & UNSAT & UNSAT   & UNSAT  &  & 32              & UNSAT  & UNSAT  & UNSAT  & SAT\\
\bottomrule
\end{tabular}
\end{table}

\begin{table}[]
\caption{Robustness of seed inputs from heart disease dataset under varying noise bounds}
\label{tab:heart}
\begin{tabular}{lrrrrl|lrrrr}
\toprule
\multicolumn{1}{c}{\multirow{2}{*}{\textbf{Input}}} & \multicolumn{4}{c}{\textbf{Result with   Noise:}}  &  & \multicolumn{1}{c}{\multirow{2}{*}{\textbf{Input}}} & \multicolumn{4}{c}{\textbf{Result with   Noise:}}   \\
\multicolumn{1}{c}{} & \multicolumn{1}{l}{$2\%$} & \multicolumn{1}{l}{$5\%$} & \multicolumn{1}{l}{$8\%$} & \multicolumn{1}{l}{$10\%$} &  & \multicolumn{1}{c}{}         & \multicolumn{1}{l}{$2\%$} & \multicolumn{1}{l}{$5\%$} & \multicolumn{1}{l}{$8\%$} & \multicolumn{1}{l}{$10\%$} \\
\midrule
1  & UNSAT & UNSAT  & UNSAT  & UNSAT  &  & 19              & UNSAT   & SAT  & SAT  & SAT    \\
2  & UNSAT & UNSAT  & UNSAT  & UNSAT   &  & 20             & UNSAT  & UNSAT  & UNSAT  & UNSAT    \\
3  & UNSAT  & UNSAT  & UNSAT  & UNSAT   &  & 21             & UNSAT  & UNSAT   & UNSAT  & UNSAT  \\
4  & UNSAT  & UNSAT & SAT & SAT  &  & 22                   & SAT  & SAT  & SAT  & SAT    \\
5  & UNSAT & UNSAT  & UNSAT & UNSAT  &  & 23               & SAT  & SAT   & SAT  & SAT \\
6 & UNSAT & UNSAT  & UNSAT & UNSAT &  & 24                 & UNSAT  & UNSAT  & UNSAT   & UNSAT   \\
7  & UNSAT  & UNSAT  & UNSAT & UNSAT  &  & 25              & UNSAT   & UNSAT  & UNSAT  & UNSAT \\
8  & UNSAT  & UNSAT & UNSAT  & UNSAT &  & 26                & UNSAT & UNSAT & UNSAT  & UNSAT \\
9 & UNSAT  & SAT  & SAT  & SAT  &  & 27                    & SAT  & SAT  & SAT  & SAT  \\
10  & UNSAT  & SAT  & SAT   & SAT  &  & 28                 & UNSAT  & UNSAT  & UNSAT & UNSAT  \\
11 & UNSAT  & UNSAT  & UNSAT & UNSAT  &  & 29              & UNSAT  & UNSAT   & UNSAT  & SAT   \\
12 & UNSAT & UNSAT  & UNSAT  & UNSAT  &  & 30               & UNSAT  & UNSAT  & UNSAT  & UNSAT   \\
13  & UNSAT & UNSAT  & UNSAT  & SAT   &  & 31              & UNSAT   & UNSAT  & UNSAT   & UNSAT   \\
14  & UNSAT & UNSAT  & UNSAT  & UNSAT  &  & 32             & UNSAT  & UNSAT & UNSAT   & UNSAT   \\
15  & SAT  & SAT  & SAT  & SAT   &  & 33                   & UNSAT & UNSAT   & UNSAT   & UNSAT  \\
16 & UNSAT & UNSAT & UNSAT   & UNSAT &  & 34                & UNSAT  & UNSAT  & UNSAT  & UNSAT  \\
17  & UNSAT  & UNSAT  & SAT & SAT  &  & 35                 & SAT   & SAT   & SAT   & SAT   \\
18  & UNSAT  & UNSAT  & UNSAT & SAT  &  & 36               & UNSAT  & UNSAT & UNSAT    & UNSAT \\
\bottomrule
\end{tabular}
\end{table}
As explained in Section \ref{sec:fannet}, the Kripke structure model for NNs generated by FANNet is quite large, owing to the enumeration of noise applied to seed inputs. In contrast, the formal model generated by the FANNet$+$ is considerably smaller due to the optimizations for state--space reduction used. Hence, the framework provides same results (i.e., SAT or UNSAT) for both robustness verification and noise tolerance determination. However, the execution times of the frameworks are significantly different. Tables \ref{tab:leu} and \ref{tab:heart} summarise the the robustness of the seed inputs from the leukemia and heart disease datasets, respectively, under the incidence of increasing noise bounds.
~\\
\begin{figure}[ht]
	\centering
	\includegraphics[width=\linewidth]{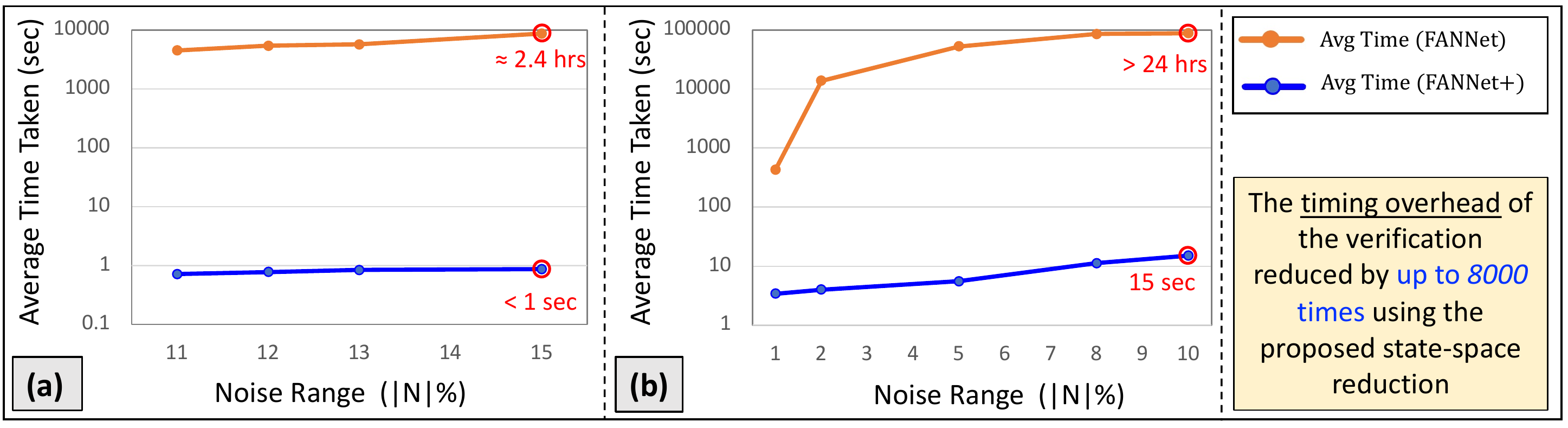}
	\caption{Comparison of Timing Overhead using FANNet and FANNet$+$: (a) plots the results obtained from the NN trained on the leukemia dataset, while (b) plots results from the NN trained on the heart disease dataset.}
	\label{FANNetvsMeth}
	\Description[Reduced timing overhead using FANNet$+$]{Even for cases where FANNet takes hours for verify for the  given input noise, FANNet$+$ is able to verify NN model, for same noise range, within a few seconds.}
\end{figure}

\noindent \textit{\textbf{FANNet versus FANNet$+$ --}} 
We run both frameworks for the NNs trained on the leukemia and the heart disease datasets. Figs. \ref{FANNetvsMeth} shows the average execution time for verifying NNs for both datasets, for seed inputs robust to the applied noise. Identical noise bounds for seed inputs were used for both frameworks. Both frameworks use iterative noise reduction, leading to noise tolerance for NNs.

As observed in Fig. \ref{bias}(a) and Fig.\ref{fannet-sens} for NN trianed on leukemia dataset, both frameworks lead to the same noise tolerance i.e., $11\%$ (also observe Table \ref{tab:leu}). The same is observed with NN trained on the heart disease dataset, having a noise tolerance of $<1\%$. However, the execution time for property verification is significantly larger for FANNet, as shown in Fig. \ref{FANNetvsMeth}. For the given NNs, FANNet$+$ provides a significant improvement over FANNet in terms of timing--cost, by reducing the timing overheard by a factor of up to $8000$ times. This makes FANNet$+$ suitable for the analysis of relatively larger NNs.

\subsection{Counterexample Analysis for detecting NN's Training Bias and Input Node Sensitivity}
\begin{figure}[ht]
	\centering
	\includegraphics[width=\linewidth]{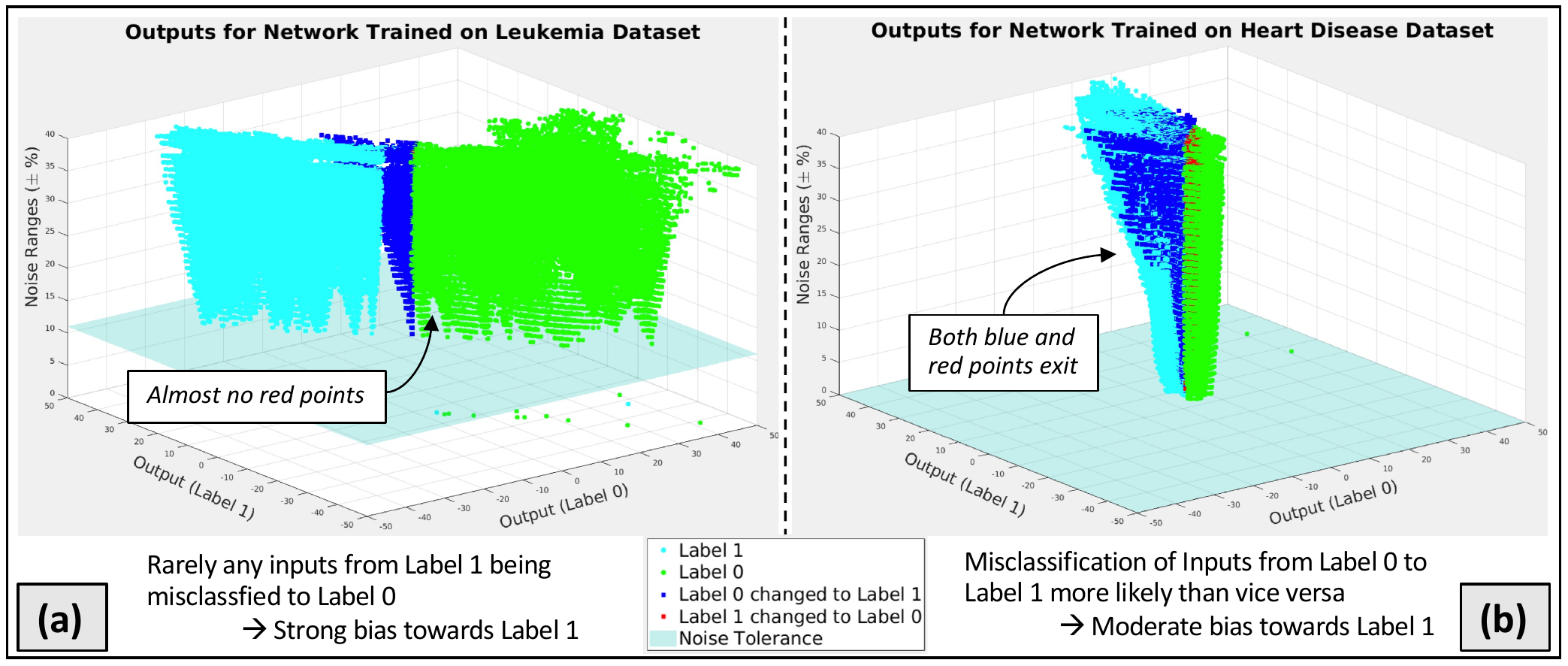}
	\caption{Output classification of NNs trained on (a) Leukemia dataset and (b) Heart disease dataset, mapped with respect to the noise applied to seed inputs, using FANNet$+$. An unequal number of red and blue points indicates a bias in trained networks.}
	\label{bias}
	\Description[Both graphs in (a) and (b) indicate a significantly large number of blue points as compared to red]{In each graph, the blue and red points indicates the inputs misclassified from each of the output classes.}	
\end{figure}

\begin{figure}[ht]
	\centering
	\includegraphics[width=\linewidth]{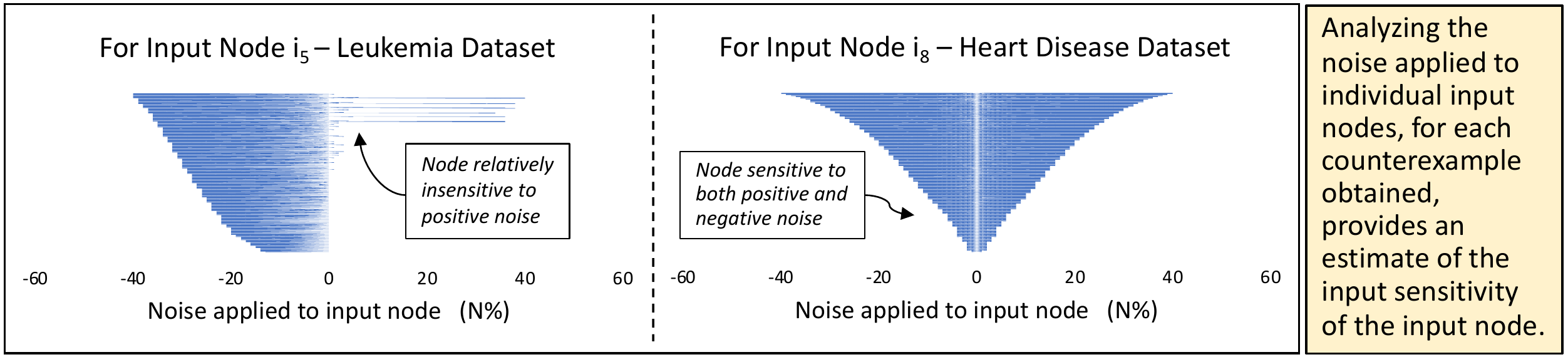}
	\caption{Plots of noise applied to individual input nodes (using FANNet$+$), that lead to misclassification.}
	\label{sens}
	\Description[]{Fully described in the text.}	
\end{figure}

With the reduction in timing overhead, it is possible to run the framework for small timeout, and yet be able to collect a large database of misclassifying noise vectors, i.e., counterexamples. Analyzing the NN outputs for these counterexamples provide insights regarding training bias and input node sensitivity, as shown in Figs. \ref{bias} and \ref{sens}.

In Fig. \ref{bias}(a), Label $1$ corresponds to ALL leukemia while Label $0$ corresponds to AML leukemia. On the other hand, Label $1$ and $0$ correspond to the cases with and without blood vessels narrowing, respectively, in Fig. \ref{bias}(b). As observed in the figures, for the NN trained on leukemia dataset, even when the large noise is applied to inputs, ALL inputs  are rarely misclassified to AML. From the description of available datasets (discussed in Section \ref{sec:nn}), it is known that training dataset for leukemia is significantly imbalanced, i.e., approximately $70\%$ inputs belong to ALL. Hence, the obtained results indicate a strong bias in the resulting trained NN likely due to the imbalance in dataset.

On the contrary, for the NN trained on heart disease dataset, outputs from both classes are misclassified even though the misclassifications from Label $0$ to Label $1$ are more likely. The training dataset for heart disease does not have the same class imbalance as the leukemia dataset. It contains inputs from the two input classes Labels $1$ and $0$ with a ratio of $55:45$. This likely accounts for the relatively moderate bias observed in the NN.

\begin{figure}[ht]
	\centering
	\includegraphics[width=\linewidth]{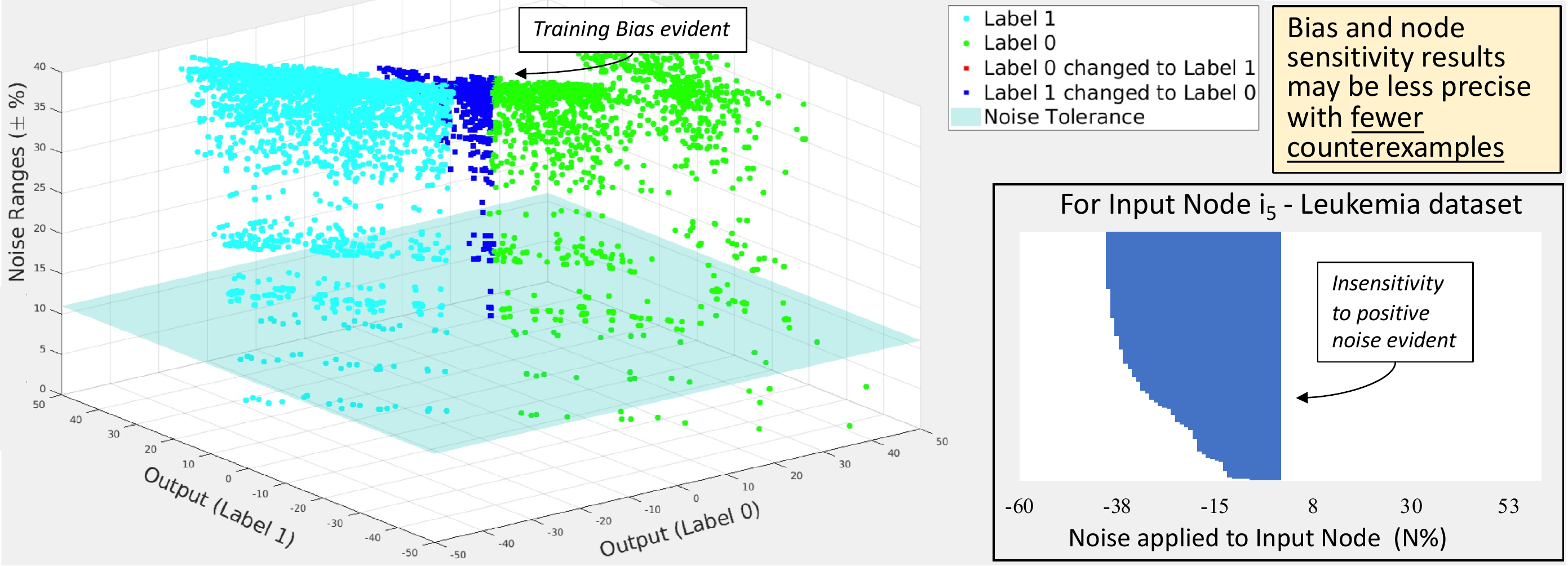}
	\caption{Training bias and input node sensitivity of the NN trained on leukemia dataset, analyzed using the counterexample database obtained using FANNet.}
	\label{fannet-sens}
	\Description[]{Fully described in the text.}	
\end{figure}

For the misclassifying noise vectors, observing noise applied to the individual input node provides insights to the sensitivity of the input nodes. For instance, input node $i_5$ from the NN trained on the leukemia dataset, shown in Fig. \ref{sens}, was observed to delineate rarely any misclassifications for the positive values of the applied noise. This suggests the node to be insensitive to positive noise, for the trained NN. However, this was not the case for any of the input nodes for the NN trained on heart disease dataset. For instance, node $i_8$ in Fig. \ref{sens} gives an example of a typical input node that is not sensitive to any specific input noise.

~\\
\noindent \textit{\textbf{FANNet versus FANNet$+$ --}} 
We performed the same experiment for NN trained on Leukemia using FANNet, as shown in Fig. \ref{fannet-sens}. As discussed earlier, due to the large timing overhead of FANNet, the counterexamples obtained compose a smaller counterexample database. This is evident in the sparse $3$D plot in Fig. \ref{fannet-sens}. Hence, even though both FANNet and FANNet$+$ indicate the NN to be biased towards Label 1 and input node $i_5$ to be insensitive to noise, the results obtained by FANNet$+$ are more precise. For instance, given the noise bounds of $40\%$, Fig. \ref{fannet-sens} indicate node $i_5$ to be completely insensitive to any positive noise. However, FANNet$+$ (as shown in Fig. \ref{sens}(a)) predicts the node to be relatively insensitive to positive noise, but may still lead to NN misclassification with certain noise patterns. Hence, the larger counterexample database with FANNet$+$ is able to provide more precise results for training bias and input node sensitivity, as compared to FANNet.

\subsection{Safety Verification} \label{subsec:safe}
For NN properties, like safety, which involve large input domain, coarse-grain verification with input step sizes for ACAS Xu NNs as shown in Table \ref{coarse}, is first used. The verification for each property, for each NN, takes only a few seconds to complete. As mentioned earlier in Section \ref{sec:pm}, coarse--grain verification is suitable for NNs and properties where large segments of input domain violate the property. This does not hold true for ACAS Xu NNs. Hence, no counterexamples to property violation were found.

\begin{table}[ht]
\caption{Input step sizes used for coarse--grain verification of ACAS Xu safety properties.}
\begin{tabular}{>{\centering\arraybackslash}m{1.6cm} >{\centering\arraybackslash}m{2.5cm} >{\centering\arraybackslash}m{2.5cm} >{\centering\arraybackslash}m{2.5cm} >{\centering\arraybackslash}m{2.5cm}}
\toprule
\textbf{Input} & \multicolumn{4}{c}{\textbf{Step size for sampling input for each safety property}} \\
 \textbf{Node} & P$1$ & P$2$ & P$3$  & P$4$       \\
\midrule
$i_1$  & $10,000$   & $10,000$  & $10^{-9}$  & $1000$     \\
$i_2$  & $1$        & $1$       & $10^{-9}$  & $10^{-1}$  \\
$i_3$  & $1$        & $1$       & $10^{-9}$  & --         \\
$i_4$  & $500$      & $500$     & $10^{-9}$  & $400$      \\
$i_5$  & $500$      & $500$     & $10^{-9}$  & $400$  \\   
\bottomrule
\end{tabular}
\label{coarse}
\end{table}
\begin{figure*}[ht]
	\centering
	\includegraphics[width=\linewidth]{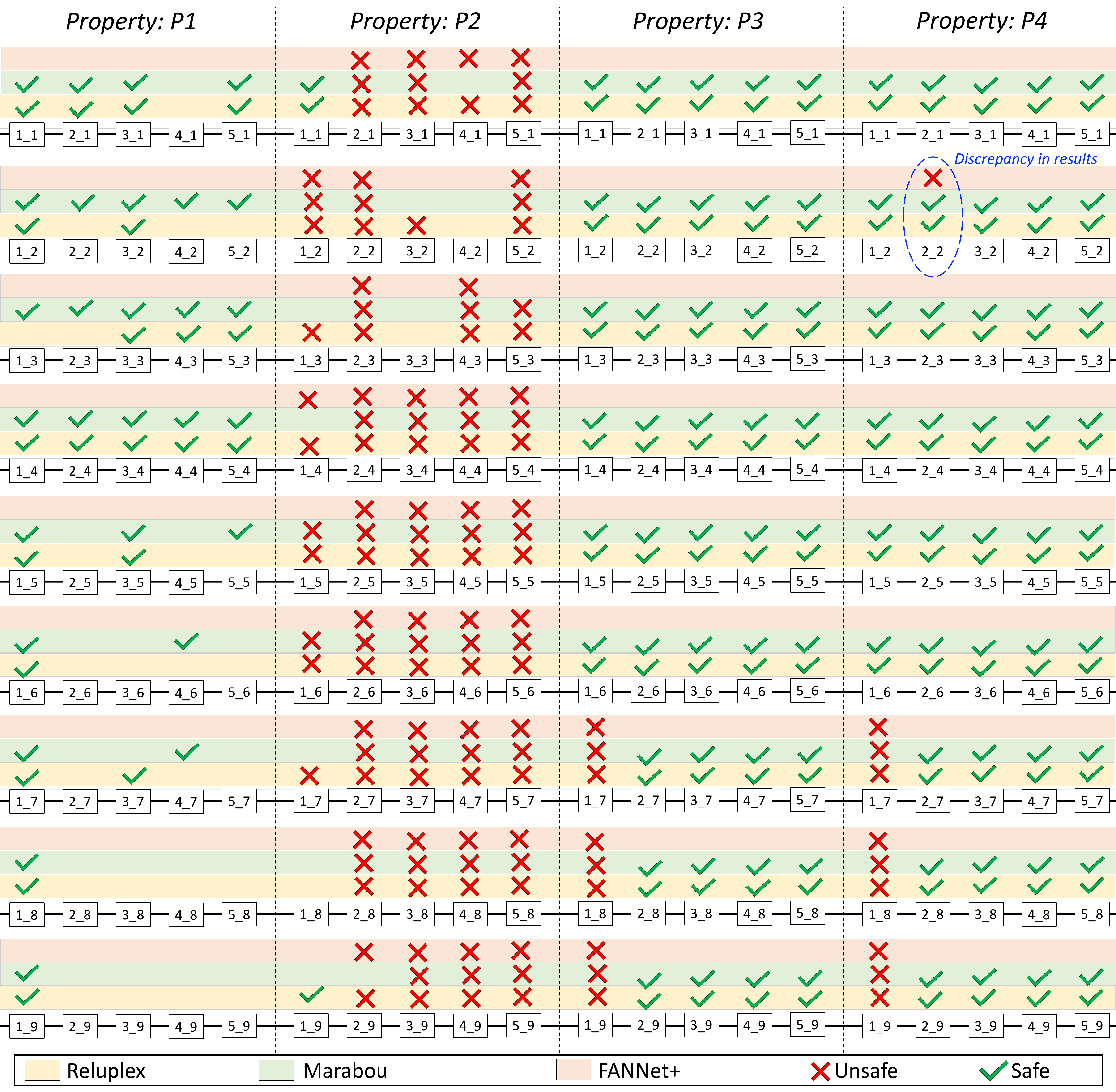}
	\caption{Safety verification using Reluplex, Marabou and FANNet$+$: the frameworks either identify whether the bounded input domain is safe or unsafe for desired network safety properties, or remain inconclusive. In case of discrepancy in result, FANNet$+$ is found to provide correct result as opposed to the the other frameworks.}
	\label{safety_comparison}
	\Description[]{Fully described in the text.}	
\end{figure*}

Next, random input segmentation is deployed for the safety verification of ACAS Xu properties. The number of segments for each input node for the safety properties is given in Table \ref{bin}. Based on the chosen \textit{variable} and \textit{fixed} sets, and the input segments of the nodes from \textit{fixed} set, verification of each NN is split into multiple smaller verification sub-problems. As stated earlier, these verification problems are independent and hence, given sufficient computation resources, they can potentially all be verified in parallel. If any of these sub-problems return a SAT, the property is said to be violated for the NN. Likewise, if any of the sub-problems times out, the entire property is considered to have timed out, since the model checker is unable to find a result for a sub-section of the input domain. Otherwise, the verification is deemed to have terminated without a solution. It must be highlighted here that the framework does not return UNSAT since the use of \textit{fixed} set introduces a certain degree of incompleteness, with respect to the input domain verified for the property. However, this notion of incompleteness is not the same as the \textit{incompleteness of the formal model} observed in numerous state--of--the--art \cite{DeepZ,matlab-incomp,DeepPoly,matlab}, which may lead to false positive. On the other hand, FANNet$+$ does not lead to false positives in the results.

\begin{table}[ht]
\caption{Number of segments of input domain corresponding to every input node, for each property under analysis.}
\begin{tabular}{>{\centering}m{3.2cm}>{\centering}m{2.1cm}>{\centering}m{2.1cm}>{\centering}m{2.1cm}>{\centering\arraybackslash}m{2.1cm}}
\toprule
\textbf{Input Node} & P$1$ & P$2$ & P$3$ & P$4$ \\
\midrule
$i_1$  & 3  & 3   & 2  & 2  \\
$i_2$  & 4  & 4   & 3  & 3  \\
$i_3$  & 4  & 4   & 2  & 1  \\
$i_4$  & 2  & 2   & 4  & 4  \\
$i_5$  & 2  & 2   & 4  & 4  \\
\bottomrule       
\end{tabular}
\label{bin}
\end{table}

The results of the safety verification are compared to those obtained from Reluplex and Marabou. For all problems that provide SAT results with Reluplex and Marabou, FANNet$+$ is also able to find the property violation unless the verification times out. These are summarized in Fig. \ref{safety_comparison}, where the success of FANNet$+$ in identifying unsafe networks is comparable to those of the state--of--the--art. It is interesting to note that FANNet$+$ was also able to find a property violation for the network $2\_2$ with property P$4$, although both Reluplex and Marabou return UNSAT for the stated property. We confirmed the validity of the obtained counterexample using \texttt{Maraboupy}. The complete list of results for the safety properties using Reluplex, Marabou and FANNet$+$ is also provided in the appendix.

~\\
\noindent \textit{\textbf{FANNet versus FANNet$+$ --}} 
As opposed to FANNet$+$, which leverages coarse--grain verification and random input segmentation to split input domain prior to verification, FANNet relies on bounds of entire input domain. Hence, the verification of safety properties of ACAS Xu NNs was infeasible with FANNet, even with a timeout of $24$ hours. 


\section{Conclusion} \label{sec:conclusion}

Formally analyzing Neural Networks (NNs) is an actively sought research domain. 
Towards this end, a model checking--based framework FANNet was introduced earlier, which could verify robustness of trained NNs to constrained noise, and also analyze NNs' noise tolerance, input node sensitivity and any underlying training bias. 
This paper proposes a more scalable model checking--based framework FANNet$+$, which improves over FANNet in multiple directions.
Firstly, it employs a state--space reduction strategy to ensure a more manageable size of the resulting state--transition system. 
Secondly, it introduced input domain segmentation approaches, including our novel random input segmentation approach, which uses a divide--and--conquer strategy to break the verification problem into smaller sub--problems. 
Thirdly, apart from the multiple NN properties mentioned earlier, the new framework FANNet$+$ extends the scope of NN analysis by adding the verification of safety properties to the framework. 
The aforementioned improvements were found to scale the framework for NN's with up to $80$ times more NN parameters, while reducing the timing overhead by a factor of up to $8000$ for the considered examples. 
The applicability of the framework was shown using multiple NNs, including the well--known ACAS Xu NNs benchmark.
\begin{acks} \label{sec:app}
This work was partially supported by Doctoral College Resilient Embedded Systems which is run jointly by TU Wien's Faculty of Informatics and FH-Technikum Wien.
\end{acks}

\bibliographystyle{ACM-Reference-Format}
\bibliography{ref}

\appendix
\LTcapwidth=\textwidth
\section{Safety Verification results using Reluplex, Marabou and FANNet\texorpdfstring{$+$}{Lg}}

This section provides the detailed results obtained after analyzing the safety properties \ref{p1}-\ref{p4} for ACAS Xu networks using Reluplex \cite{reluplex}, Marabou \cite{marabou} and FANNet$+$. A consistent timeout of two-hours was used for all the experiments. As can be seen from the results in Tables \ref{tab:app:p1p2} and \ref{tab:app:p3p4}, for all experiments, whenever the renowned tools Reluplex and/or Marabou find satisfying solution to the violation of property, FANNet$+$ is also able to find the counterexample(s) within the pre-defined timeout. 

\setcounter{table}{0}
\renewcommand\thetable{A.\arabic{table}}

\begin{ThreePartTable}
\begin{TableNotes}
\footnotesize
\item SAT: satisfiable solution found \tab~~~~~~~~~~~~ UNSAT: property unsatisfiable
\item TO: timeout (execution time exceeded $2$ hours) \tab~~~~~ NF: no counterexample found
\end{TableNotes}

\begin{longtable}{>{\centering}m{1.8cm}>{\centering}m{1.5cm}>{\centering}m{1.5cm}>{\centering}m{1.5cm}>{\centering}m{1.5cm}>{\centering}m{1.5cm}>{\centering\arraybackslash}m{1.5cm}}

\caption{Results of verifying properties P$1$ and P$2$ for ACAS Xu Neural Networks} \label{tab:app:p1p2} \\
\toprule
 \multirow{2}{*}{\textbf{Network}} & \multicolumn{2}{c}{\textbf{Reluplex}} & \multicolumn{2}{c}{\textbf{Marabou}} & \multicolumn{2}{c}{\textbf{FANNet$+$}} \\
   & P$1$  & P$2$  & P$1$  & P$2$  & P$1$  & P$2$  \\
\midrule
\endfirsthead

\toprule
 \multirow{2}{*}{\textbf{Network}} & \multicolumn{2}{c}{\textbf{Reluplex}} & \multicolumn{2}{c}{\textbf{Marabou}} & \multicolumn{2}{c}{\textbf{FANNet$+$}} \\
   & P$1$  & P$2$  & P$1$  & P$2$  & P$1$  & P$2$  \\
\midrule
\endhead

\bottomrule
\endfoot

\bottomrule
\insertTableNotes \\  

\endlastfoot

$1\_1$  & UNSAT & UNSAT & UNSAT & UNSAT & TO & TO   \\
$1\_2$  & UNSAT & SAT   & UNSAT & SAT   & TO & SAT  \\
$1\_3$  & TO    & SAT   & UNSAT & TO    & TO & TO   \\
$1\_4$  & UNSAT & SAT   & UNSAT & TO    & TO & SAT  \\
$1\_5$  & UNSAT & SAT   & UNSAT & SAT   & TO & TO   \\
$1\_6$  & UNSAT & SAT   & UNSAT & SAT   & TO & TO   \\
$1\_7$  & UNSAT & SAT   & UNSAT & TO    & TO & TO  \\
$1\_8$  & UNSAT & TO    & UNSAT & TO    & TO & TO   \\
$1\_9$  & UNSAT & UNSAT & UNSAT & TO    & TO & TO  \\
$2\_1$  & UNSAT & SAT   & UNSAT & SAT   & TO & SAT  \\
$2\_2$  & TO    & SAT   & UNSAT & SAT   & TO & SAT  \\
$2\_3$  & TO    & SAT   & UNSAT & SAT   & TO & SAT  \\
$2\_4$  & UNSAT & SAT   & UNSAT & SAT   & TO & SAT  \\
$2\_5$  & TO    & SAT   & TO    & SAT   & TO & SAT  \\
$2\_6$  & TO    & SAT   & TO    & SAT   & TO & SAT  \\
$2\_7$  & TO    & SAT   & TO    & SAT   & TO & SAT  \\
$2\_8$  & TO    & SAT   & TO    & SAT   & TO & SAT  \\
$2\_9$  & TO    & SAT   & TO    & TO    & TO & SAT  \\
$3\_1$  & UNSAT & SAT   & UNSAT & SAT   & TO & SAT  \\
$3\_2$  & UNSAT & SAT   & UNSAT & TO    & TO & TO   \\
$3\_3$  & UNSAT & TO    & UNSAT & TO    & TO & TO   \\
$3\_4$  & UNSAT & SAT   & UNSAT & SAT   & TO & SAT  \\
$3\_5$  & UNSAT & SAT   & UNSAT & SAT   & TO & SAT  \\
$3\_6$  & TO    & SAT   & TO    & SAT   & TO & SAT  \\
$3\_7$  & UNSAT & SAT   & TO    & SAT   & TO & SAT  \\
$3\_8$  & TO    & SAT   & TO    & SAT   & TO & SAT  \\
$3\_9$  & TO    & SAT   & TO    & SAT   & TO & SAT  \\
$4\_1$  & TO    & SAT   & TO    & TO    & TO & SAT  \\
$4\_2$  & TO    & TO    & UNSAT & TO    & TO & TO   \\
$4\_3$  & UNSAT & SAT   & UNSAT & SAT   & TO & SAT  \\
$4\_4$  & UNSAT & SAT   & UNSAT & SAT   & TO & SAT  \\
$4\_5$  & TO    & SAT   & TO    & SAT   & TO & SAT  \\
$4\_6$  & TO    & SAT   & UNSAT & SAT   & TO & SAT  \\
$4\_7$  & TO    & SAT   & UNSAT & SAT   & TO & SAT  \\
$4\_8$  & TO    & SAT   & TO    & SAT   & TO & SAT  \\
$4\_9$  & TO    & SAT   & TO    & SAT   & TO & SAT  \\
$5\_1$  & UNSAT & SAT   & UNSAT & SAT   & TO & SAT  \\
$5\_2$  & TO    & SAT   & UNSAT & SAT   & TO & SAT  \\
$5\_3$  & UNSAT & SAT   & UNSAT & SAT   & TO & TO   \\
$5\_4$  & UNSAT & SAT   & UNSAT & SAT   & TO & SAT  \\
$5\_5$  & TO    & SAT   & UNSAT & SAT   & TO & SAT  \\
$5\_6$  & TO    & SAT   & TO    & SAT   & TO & SAT  \\
$5\_7$  & TO    & SAT   & TO    & SAT   & TO & SAT  \\
$5\_8$  & TO    & SAT   & TO    & SAT   & TO & SAT  \\
$5\_9$  & TO    & SAT   & TO    & SAT   & TO & SAT  \\

\end{longtable}
\end{ThreePartTable}

\begin{ThreePartTable}
\begin{TableNotes}
\footnotesize
\item SAT: satisfiable solution found \tab~~~~~~~~~~~~ UNSAT: property unsatisfiable
\item TO: timeout (execution time exceeded $2$ hours) \tab ~~~~~ NF: no counterexample found

\end{TableNotes}
\begin{longtable}{>{\centering}m{1.8cm}>{\centering}m{1.5cm}>{\centering}m{1.5cm}>{\centering}m{1.5cm}>{\centering}m{1.5cm}>{\centering}m{1.5cm}>{\centering\arraybackslash}m{1.5cm}}

\caption{Results of verifying properties P$3$ and P$4$ for ACAS Xu Neural Networks} \label{tab:app:p3p4}\\
\toprule
\multirow{2}{*}{\textbf{Network}} & \multicolumn{2}{c}{\textbf{Reluplex}} & \multicolumn{2}{c}{\textbf{Marabou}} & \multicolumn{2}{c}{\textbf{FANNet$+$}} \\
& P$3$   & P$4$  & P$3$  & P$4$   & P$3$   & P$4$ \\
\midrule
\endfirsthead

\toprule
 \multirow{2}{*}{\textbf{Network}} & \multicolumn{2}{c}{\textbf{Reluplex}} & \multicolumn{2}{c}{\textbf{Marabou}} & \multicolumn{2}{c}{\textbf{FANNet$+$}} \\
& P$3$   & P$4$  & P$3$  & P$4$   & P$3$   & P$4$ \\
\midrule
\endhead

\bottomrule
\endfoot

\bottomrule
\insertTableNotes \\  
\endlastfoot

$1\_1$ & UNSAT  & UNSAT  & UNSAT & UNSAT  & TO  & NF  \\
$1\_2$ & UNSAT  & UNSAT  & UNSAT & UNSAT  & TO  & TO  \\
$1\_3$ & UNSAT  & UNSAT  & UNSAT & UNSAT  & TO  & TO  \\
$1\_4$ & UNSAT  & UNSAT  & UNSAT & UNSAT  & TO  & NF  \\
$1\_5$ & UNSAT  & UNSAT  & UNSAT & UNSAT  & TO  & NF  \\
$1\_6$ & UNSAT  & UNSAT  & UNSAT & UNSAT  & TO  & TO  \\
$1\_7$ & SAT    & SAT    & SAT   & SAT    & SAT & SAT \\
$1\_8$ & SAT    & SAT    & SAT   & SAT    & SAT & SAT \\
$1\_9$ & SAT    & SAT    & SAT   & SAT    & SAT & SAT \\
$2\_1$ & UNSAT  & UNSAT  & UNSAT & UNSAT  & TO  & TO  \\
$2\_2$ & UNSAT  & UNSAT  & UNSAT & UNSAT  & TO  & SAT \\
$2\_3$ & UNSAT  & UNSAT  & UNSAT & UNSAT  & TO  & NF  \\
$2\_4$ & UNSAT  & UNSAT  & UNSAT & UNSAT  & TO  & NF  \\
$2\_5$ & UNSAT  & UNSAT  & UNSAT & UNSAT  & TO  & TO  \\
$2\_6$ & UNSAT  & UNSAT  & UNSAT & UNSAT  & TO  & TO  \\
$2\_7$ & UNSAT  & UNSAT  & UNSAT & UNSAT  & TO  & TO  \\
$2\_8$ & UNSAT  & UNSAT  & UNSAT & UNSAT  & NF  & TO  \\
$2\_9$ & UNSAT  & UNSAT  & UNSAT & UNSAT  & TO  & NF  \\
$3\_1$ & UNSAT  & UNSAT  & UNSAT & UNSAT  & TO  & NF  \\
$3\_2$ & UNSAT  & UNSAT  & UNSAT & UNSAT  & TO  & NF  \\
$3\_3$ & UNSAT  & UNSAT  & UNSAT & UNSAT  & TO  & NF  \\
$3\_4$ & UNSAT  & UNSAT  & UNSAT & UNSAT  & TO  & NF  \\
$3\_5$ & UNSAT  & UNSAT  & UNSAT & UNSAT  & TO  & TO  \\
$3\_6$ & UNSAT  & UNSAT  & UNSAT & UNSAT  & TO  & NF  \\
$3\_7$ & UNSAT  & UNSAT  & UNSAT & UNSAT  & TO  & NF  \\
$3\_8$ & UNSAT  & UNSAT  & UNSAT & UNSAT  & TO  & NF  \\
$3\_9$ & UNSAT  & UNSAT  & UNSAT & UNSAT  & TO  & NF  \\
$4\_1$ & UNSAT  & UNSAT  & UNSAT & UNSAT  & TO  & NF  \\
$4\_2$ & UNSAT  & UNSAT  & UNSAT & UNSAT  & TO  & NF  \\
$4\_3$ & UNSAT  & UNSAT  & UNSAT & UNSAT  & TO  & NF  \\
$4\_4$ & UNSAT  & UNSAT  & UNSAT & UNSAT  & TO  & NF  \\
$4\_5$ & UNSAT  & UNSAT  & UNSAT & UNSAT  & TO  & NF  \\
$4\_6$ & UNSAT  & UNSAT  & UNSAT & UNSAT  & TO  & NF  \\
$4\_7$ & UNSAT  & UNSAT  & UNSAT & UNSAT  & TO  & NF  \\
$4\_8$ & UNSAT  & UNSAT  & UNSAT & UNSAT  & TO  & TO  \\
$4\_9$ & UNSAT  & UNSAT  & UNSAT & UNSAT  & TO  & NF  \\
$5\_1$ & UNSAT  & UNSAT  & UNSAT & UNSAT  & TO  & NF  \\
$5\_2$ & UNSAT  & UNSAT  & UNSAT & UNSAT  & NF  & NF  \\
$5\_3$ & UNSAT  & UNSAT  & UNSAT & UNSAT  & TO  & NF  \\
$5\_4$ & UNSAT  & UNSAT  & UNSAT & UNSAT  & TO  & NF  \\
$5\_5$ & UNSAT  & UNSAT  & UNSAT & UNSAT  & TO  & NF  \\
$5\_6$ & UNSAT  & UNSAT  & UNSAT & UNSAT  & TO  & NF  \\
$5\_7$ & UNSAT  & UNSAT  & UNSAT & UNSAT  & TO  & NF  \\
$5\_8$ & UNSAT  & UNSAT  & UNSAT & UNSAT  & TO  & NF  \\
$5\_9$ & UNSAT  & UNSAT  & UNSAT & UNSAT  & TO  & NF
\\

\end{longtable}
\end{ThreePartTable}

Interestingly, FANNet$+$ is also able to find violation to property \ref{p4} for the network $2\_2$, i.e., the experiment for which both Reluplex and Marabou return UNSAT. The satisfying input leading to violation of the property, as found by FANNet$+$ is as follows:
$[1600, 3, 0, 1100, 720]$
\end{document}